\documentclass[nonblindrev]{informs3a}

\input{latexmacro}
\usepackage{algorithm,algorithmic}
\usepackage{diagbox}
\usepackage[dvipsnames]{xcolor}
\usepackage{booktabs}
\usepackage{hyperref}
\usepackage{float}

\usepackage{subfigure}
\usepackage{caption}
\newlength\figureheight 
\newlength\figurewidth 
\newcommand{\killpunct}[1]{}

\OneAndAHalfSpacedXI

\usepackage{natbib}
 \bibpunct[, ]{(}{)}{,}{a}{}{,}%
 \def\bibfont{\small}%
\TheoremsNumberedThrough     
\ECRepeatTheorems

\EquationsNumberedThrough    

\MANUSCRIPTNO{MS-0001-1922.65}

\begin{document}



\RUNAUTHOR{}

\RUNTITLE{Continuous cGAN with Generator Regularization}

\TITLE{Continuous Conditional Generative Adversarial Networks (cGAN) with Generator Regularization}

\ARTICLEAUTHORS{%
\AUTHOR{Yufeng Zheng\thanks{The authors contribute equally.} \qquad Yunkai Zhang\footnotemark[1] \qquad Zeyu Zheng} 

\medskip

\AFF{Department of Industrial Engineering and Operations Research, University of California, Berkeley, CA 94720\\ \EMAIL{yufeng\_zheng, yunkai\_zhang, zyzheng@berkeley.edu}} 
\smallskip
\AUTHOR{Draft Version: Mar 27, 2021}
} 

\ABSTRACT{Conditional Generative Adversarial Networks are known to be difficult to train, especially when the conditions are continuous and high-dimensional. To partially alleviate this difficulty, we propose a simple generator regularization term on the GAN generator loss in the form of Lipschitz penalty. Thus, when the generator is fed with neighboring conditions in the continuous space, the regularization term will leverage the neighbor information and push the generator to generate samples that have similar conditional distributions for each neighboring condition. We analyze the effect of the proposed regularization term and demonstrate its robust performance on a range of synthetic and real-world tasks. 
}%

\maketitle
\section{Introduction}

Conditional Generative Adversarial Networks (cGANs) \cite{mirza2014conditional} are a powerful class of generative models where the goal is to learn a mapping from input to output distributions conditioned on some auxiliary information, such as class labels \cite{mirza2014conditional, miyato2018cgans}, images \cite{wang2018highresolution, IizukaSIGGRAPH2017}, or text \cite{reed2016generative, qiao2019mirrorgan}. While cGANs have demonstrated outstanding capabilities in a wide range of conditional generation tasks, they are also known to be difficult to train since the optimization objective is cast as a min-max game between the generator network and the discriminator network. Much past work has been devoted to stabilize the training of GANs. For example, \citet{arjovsky2017wasserstein} introduces Wasserstein-GAN (WGAN) that uses the Earth Mover distance as a more explicit measure of the distribution divergence in the loss function. To better enforce the $k$-Lipschitz assumption in WGANs, \citet{gulrajani2017improved} presents a regularization term on the discriminator. \citet{yang2019diversitysensitive} studies the issue of mode-collapse, where only a small subset of the true output distribution is learned by the generator \cite{salimans2016improved}, by encouraging the generator to produce diverse outputs based on the latent input noise. On the other hand, \citet{zhang2020consistency} proposes to penalize the discriminator from being overly sensitive to small perturbations to the inputs through consistency regularization by augmenting the data passed into the discriminator during training.

However, new challenges arise when the given conditions are continuous (termed regression labels) and multi-dimensional, which are often observed in real-life scenarios. One example is to generate spatial distributions of taxi's drop-off locations conditioned on its pick-up time and locations \cite{dutordoir2018gaussian}. Another example is to synthesize facial images conditioned on age \cite{ding2021ccgan}. A common practice is to treat each distinct age as a separate class, ignoring inter-class correlations (e.g. the intrinsic similarities between age groups that are closer to each other). Additionally, if not every possible condition is represented in the training data which we denote as \textit{gaps}, the neural network generator might extend poorly to those unseen conditions. To address such concerns, \citet{ding2021ccgan} introduces CcGAN and suggests to add Gaussian noises to each sample of the input conditions in order to cover the gaps, at the cost of less sensitivity of the generator to more granular changes in the input conditions. In light of these observations, we propose a simple but effective generator regularization term on the GAN generator loss in the form of Lipschitz penalty. The intuition is that when a small perturbation is applied to any condition in the condition space, the output semantics should only change minimally. In summary, our contributions are three-fold:
\begin{itemize}
    \item Through synthetic experiments, we demonstrate CcGAN and vanilla cGANs might suffer from undesired behaviors, especially when the dimension of the given condition or the number of gaps in the training set increases.
    \item We propose a regularization approach that encourages the generator to leverage neighboring conditions in the continuous space through Lipschitz regularization without sacrificing the generator's faithfulness to the input conditions.
    \item Instead of directly penalizing the gradients at observed conditions in the training set, we regularize the gradients along the interpolations of condition pairs, effectively closing the gaps in the training set.
\end{itemize}

\section{Method}
\textbf{Problem Formulation.} Let $\mathcal{X}\subset \mathbb{R}^m, \mathcal{Y}\subset \mathbb{R}^n, \mathcal{Z}\subset \mathbb{R}^l$ be the condition space, the output space, and the latent space respectively. Denote the underlying joint distribution for $\boldsymbol{x}\in \mathcal{X}$ and $\boldsymbol{y}\in \mathcal{Y}$ as $p_r(\boldsymbol{x},\boldsymbol{y})$. Thus, the conditional distribution of $\boldsymbol{y}$ given $\boldsymbol{x}$ becomes $p_r(\boldsymbol{y}|\boldsymbol{x})$. The training set consists of $N$ observed $(\boldsymbol{x},\boldsymbol{y})$ pairs, denoted as $\{(\boldsymbol{x}_i,\boldsymbol{y}_i)\}_{i=1}^N$. Following the vanilla cGAN \cite{mirza2014conditional}, we introduce a random noise $\boldsymbol{z}\in \mathcal{Z}$ and $\boldsymbol{z}\sim p_z(\boldsymbol{z})$, where $p_z(\boldsymbol{z})$ is a predetermined easy-to-sample distribution. The goal is to train a conditional generator $G: \mathcal{X} \times \mathcal{Z} \rightarrow \mathcal{Y}$, whose inputs are the condition $\boldsymbol{x}$ and latent noise $\boldsymbol{z}$, in order to imitate the conditional distribution $p_r(\boldsymbol{y}|\boldsymbol{x})$. Our proposed gradient penalty term is suitable for most variants of cGAN losses, such as the vanilla cGAN loss \cite{mirza2014conditional}, the Wasserstein loss \cite{arjovsky2017wasserstein}, and the hinge loss \cite{miyato2018spectral}. Without loss of generality, here we illustrate the gradient penalty on the vanilla cGAN loss, where the conditional generator $G$ and discriminator $D$ are learned by jointly optimizing the following objective:
\begin{equation}\label{eq: cGAN}
\begin{aligned}
&\min_{G} \max _{D} \mathcal{L}_{cGAN}(D, G)\\
=&\mathbb{E}_{(\boldsymbol{x},\boldsymbol{y})\sim \hat{p}(\boldsymbol{x},\boldsymbol{y})}[\log D(\boldsymbol{x},\boldsymbol{y})]\\
&+\mathbb{E}_{\boldsymbol{z} \sim p_{z}(\boldsymbol{z}),\boldsymbol{x}\sim\hat{p}(\boldsymbol{x})}[\log (1-D(\boldsymbol{x}, G(\boldsymbol{x},\boldsymbol{z})))],
\end{aligned}
\end{equation}
where $\hat{p}(\boldsymbol{x},\boldsymbol{y})$ is the empirical distribution of $\{(\boldsymbol{x}_i,\boldsymbol{y}_i)\}_{i=1}^N$, and  $\hat{p}(\boldsymbol{x})$ is the empirical distribution of $\{\boldsymbol{x}_i\}_{i=1}^N$.

\textbf{Challenges of Continuous, Multi-Dimensional Conditions.} Under the given setting, cGAN commonly suffers from two problems. \textbf{(P1)} Since the condition space $\mathcal{X}$ is continuous and multi-dimensional, $\boldsymbol{x}_i$’s are very likely to be different from each other. Furthermore, at each forward propagation only one noise $\boldsymbol{z}_j$ is sampled from $p_z(\boldsymbol{z})$. Therefore, for a certain $\boldsymbol{x}_i$, the discriminator can only get the information of $G(\boldsymbol{x}_i, \boldsymbol{z}_j)$ and may find it particularly challenging to generalize to the general distribution of $G(\boldsymbol{x}_i, \mathcal{Z})$. \textbf{(P2)} As we increase the number of dimensions for $\mathcal{X}$, the conditions observed $\{\boldsymbol{x}_i\}_{i=1}^N$ become more sparse and more gaps are created. For most conditions $\boldsymbol{x}\in \mathcal{X}$, few or even no samples can be observed during training. In most cGAN literature, when training cGANs, the generator are only given and trained on the conditions observed in the training set. As a result, the generator might extend poorly when given a new condition that has never been observed in the training set.

\textbf{Generator Regularization.} To address the aforementioned issues, we propose a novel regularization of the generator and name the resulting model as \textbf{Generator Regularized-cGAN (GR-cGAN)}. We first present the expression of the regularization term, and then discuss how it can remedy these problems.

The generator regularization is based on a continuity assumption of the conditional distribution $p_r(\boldsymbol{y}|\boldsymbol{x})$. For a wide range of applications but not all, it is natural to assume that a minor perturbation to the condition $\boldsymbol{x}$ will only slightly disturb the conditional distribution $p_r(\boldsymbol{y}|\boldsymbol{x})$. On a high level, we hope that the distribution of $G(\boldsymbol{x},\boldsymbol{z})$ shifts smoothly as we change $\boldsymbol{x}$. Since directly regularizing the generator from a distribution perspective can be challenging, we instead regularize the gradient of $G(\boldsymbol{x}, \boldsymbol{z})$ with respect to $\boldsymbol{x}$. Specifically, we add the following regularization term to the generator loss, to encourage the optimized generator $G$ to minimize on this regularization term.
\begin{equation}\label{eq: gp_original}
    \mathcal{L}_{GR}(G)=\mathbb{E}_{\substack{\boldsymbol{z}\sim q(\boldsymbol{z}),\\ \boldsymbol{x}\sim \tilde{p}(\boldsymbol{x})}}||\nabla_{\boldsymbol{x}} G(\boldsymbol{x},\boldsymbol{z})||,
\end{equation}
where $\nabla_{\boldsymbol{x}} G(\boldsymbol{x},\boldsymbol{z})$ is the Jacobian matrix given by
$$
\nabla_{\boldsymbol{x}} G(\boldsymbol{x},\boldsymbol{z}) = \left[\begin{array}{ccc}
\frac{\partial G_{1}(\boldsymbol{x}, \boldsymbol{z})}{\partial x_{1}} & \cdots & \frac{\partial G_{1}(\boldsymbol{x}, \boldsymbol{z})}{\partial x_{n}} \\
\vdots & \ddots & \vdots \\
\frac{\partial G_{m}(\boldsymbol{x}, \boldsymbol{z})}{\partial x_{1}} & \cdots & \frac{\partial G_{m}(\boldsymbol{x}, \boldsymbol{z})}{\partial x_{n}}
\end{array}\right].
$$

The distribution $\tilde{p}(\boldsymbol{x})$ indicates the locations where we regularize the Jacobian matrix $\nabla_{\boldsymbol{x}} G(\boldsymbol{x},\boldsymbol{z})$, and is implicitly defined by sampling uniformly along straight lines between pairs of conditions sampled from the training set. This sampling method allows us not only to perform regularization on the conditions observed in the training set, but also to perform regularization on the conditions that have not been observed. If the conditions are nonlinear in a too complicated space, we can also project the conditions onto another vector space - for example, onto the latent space of variational autoencoders (VAEs) - before interpolations \citep{arvanitidis2018latent, chen2018metrics}. See supplementary materials for the detailed algorithm to train GR-cGAN in practice. A natural choice of the norm in Equation ($\ref{eq: gp_original}$) is a Frobenius norm, computed by
$$
||\nabla_{\boldsymbol{x}} G(\boldsymbol{x},\boldsymbol{z})||=\sqrt{\sum_{i=1}^n\sum_{j=1}^m \left[\frac{\partial G_i(\boldsymbol{x},\boldsymbol{z})}{\partial x_j}\right]^2}.
$$
Intuitively, when $\mathcal{L}_{GR}(G)$ takes a small value, for any fixed $\boldsymbol{z} = \boldsymbol{z}_0$, the output of the generator $G(\boldsymbol{x},\boldsymbol{z}_0)$ will only shift moderately and continuously as $\boldsymbol{x}$ changes.

However, the direct evaluation of Equation ($\ref{eq: gp_original}$) is computationally prohibitive when the dimensions $m$ and $n$ are high. When the dimension of the condition and the dimension of generator output are high (say, more than 100), we provide an alternative to Equation ($\ref{eq: gp_original}$) by locally approximating the gradient in a finite difference fashion:
\begin{equation}\label{eq: approx_gp}
\begin{aligned}
& \mathcal{L}_{\widetilde{GR}}(G) 
&= \mathbb{E}_{\substack{\boldsymbol{z}\sim p_z(\boldsymbol{z}),\\ \boldsymbol{x}\sim \tilde{p}(\boldsymbol{x})}} [\min(f(\boldsymbol{x}, \Delta \boldsymbol{x}, \boldsymbol{z}), \tau_1)]
\end{aligned}
\end{equation}
where
$$
f(\boldsymbol{x}, \Delta \boldsymbol{x}, \boldsymbol{z}) = \frac{||G(\boldsymbol{x} + \Delta\boldsymbol{x}, \boldsymbol{z}) - G(\boldsymbol{x}, \boldsymbol{z})||}{||\Delta \boldsymbol{x}||},
$$
$\Delta \boldsymbol{x} \sim p_{\Delta\boldsymbol{x}}(\Delta\boldsymbol{x})$ is a small perturbation added to $\boldsymbol{x}$ and $p_{\Delta \boldsymbol{x}}(\Delta\boldsymbol{x})$ is the distribution of $\Delta\boldsymbol{x}$. The distribution $p_{\Delta \boldsymbol{x}}(\Delta\boldsymbol{x})$ is designed to be a distribution centered close to zero and has a small variance, such as a normal distribution. $\tau_1$ is a bound for ensuring numerical stability. We also impose a lower bound $\tau_2$ on $\Delta\boldsymbol{x}$ for the same reason.

Finally, the cGAN objective with generator regularization now becomes
\begin{align*}\label{eq: new_cgan}
\min_{G} \max_{D} & \quad V(D, G)\\
&=\mathbb{E}_{(\boldsymbol{x},\boldsymbol{y})\sim \hat{p}(\boldsymbol{x},\boldsymbol{y})}[\log D(\boldsymbol{x},\boldsymbol{y})]\\
&\quad+\mathbb{E}_{\boldsymbol{z} \sim p_{z}(\boldsymbol{z}),\boldsymbol{x}\sim\hat{p}(\boldsymbol{x})}[\log (1-D(G(\boldsymbol{x},\boldsymbol{z})))]\\
&\quad +\lambda\cdot\mathcal{L}_{GR}(G),
\end{align*}
where $\mathcal{L}_{GR}(G)$ can be replaced by $\mathcal{L}_{\widetilde{GR}}(G)$ if we use the approximated generator regularization given by Equation ($\ref{eq: approx_gp}$). The term $\lambda$ controls the degree of regularization. In other words, a larger $\lambda$ discourages the model from reacting rapidly to small perturbations in the input conditions.

%


\textbf{How does generator regularization overcome (P1) and (P2)?}

For (P1), when cGANs are trained, a batch of $(\boldsymbol{x}_i, \boldsymbol{y}_i)$ pairs from the training set. For any $\boldsymbol{x}_i$ from this batch of data, when the generator regularization is applied, the samples in the vicinity of $\boldsymbol{x}_i$ are encouraged to facilitate the training of the generator and the discriminator. In the case where the generated distribution of $G(\boldsymbol{x}_i, \boldsymbol{z})$ with $\boldsymbol{z}\sim p_z(\boldsymbol{z})$ is concentrated on a pathological mode collapse distribution (in other words, the generator always gives almost the same distribution around a wide neighborhood of $\boldsymbol{x}_i$), the discriminator can better detect local mode collapse and learn to classify such pathological distribution as fake, thus improving the generator in return.

For (P2), when given a new condition $\boldsymbol{x}_0$ that does not exist in the train set, the conditional distribution given by the generator in GR-cGAN on $\boldsymbol{x}_0$ is similar to the conditional distribution given on the conditions in the vicinity of $\boldsymbol{x}_0$ in the training set.
If we penalize the gradient from being too large, we are effectively encouraging the model to learn a smooth transition between each pair of samples from the training set and thus generalize to close these gaps.

\textbf{Comparison with Related Work}
Many papers have denoted to resolving the mode-collapse phenomenon, such as by incoporating divergence measure to reshape the discriminator landscape \citep{gulrajani2017improved, yang2019diversitysensitive} or generating multi-modal images \citep{Huang_2018_ECCV, zhu2018multimodal}. A lot of methods focus on the relationship of GANs with changes in the latent noise or the generator architecture, but the connection between small perturbations in the conditions are relatively less studied. Notably, CcGAN \citep{ding2021ccgan} attempts to address the continuous condition issues by adding Gaussian noises to the input conditions. This implies that the models might loss granular information about the precise information of the conditions, resulting in outputs that might be less faithful to the input conditions. In particular, when there are large gaps in the dataset, CcGAN must choose large standard deviations for Gaussian noises in order to cover these gaps, which further exacerbates the issue. On the other hand, our proposed method relies on encouraging gradual changes of the output with respect to the input conditions, which does not cause the model to lose detailed information of the conditions. More experiments can be found in the Supplementary Materials.

\section{Analysis of the Proposed Regularization}
\label{sec:analysis}
We now analyze the proposed regularization term in finer details. Following the definition of Lipschitz continuity for functions, we first deliver a formal definition of continuous conditional distribution named \textbf{Lipschitz continuous conditional distribution}. Next, we present the connection between the generator regularization and Lipschitz continuous conditional distribution.

The definition of $K$-Lipschitz Continuous Conditional Distribution is given as follows.
\begin{definition}[$K$-Lipschitz Continuous Conditional Distribution]
Let $X$ and $Y$ be random variables with support $R_X$ and $R_Y$ respectively. Denote the distribution induced by $X\mid Y=y$ as $\mathcal{F}_y$. We say $X$ has a $K$-Lipschitz continuous conditional distribution with respect to $Y$, if for all $y_1, y_2\in R_Y$, the Wasserstein distance between $\mathcal{F}_{y_1}$ and $\mathcal{F}_{y_2}$ satisfies
$$
W(\mathcal{F}_{y_1},\mathcal{F}_{y_2})\le K\cdot\|y_1-y_2\|,
$$
where $W(\cdot,\cdot)$ denotes the Wasserstein distance between two distributions, and $||\cdot||$ indicates a norm.
\end{definition}

Note that when the Wasserstein distance is used to evaluate the distance between two probability distributions, the cGANs can be extended to conditional Wasserstein GANs. Other distances to quantify the gap between two conditional distributions can also be adapted. 


Given two arbitrary conditions $\boldsymbol{x}_1$ and $\boldsymbol{x}_2$, suppose that the generator satisfies
$$
||G(x_1, z_0) - G(x_2, z_0)|| \le K_0\cdot ||x_1-x_2||
$$
for any $\boldsymbol{z}_0$. The conditional distribution given by the generator on $\boldsymbol{x}_1$ and $\boldsymbol{x}_2$ are $G(\boldsymbol{x}_1, \boldsymbol{z})$ and $G(\boldsymbol{x}_2,\boldsymbol{z})$ with $\boldsymbol{z}\sim p_z(\boldsymbol{z})$ respectively. Notice that when the generator regularization is applied, the term $K_0$ will be pushed to a smaller level. It is therefore evident that
$$
W(G(\boldsymbol{x}_1,\boldsymbol{ z}), G(\boldsymbol{x}_2, \boldsymbol{z}))\le K_0\cdot ||x_1-x_2||,
$$
which indicates the conditional distribution learned by the generator is a $K_0$-Lipschitz continuous conditional distribution with respect to $\boldsymbol{x}$. The proof steps are given in supplementary material. With the use of generator regularization, the conditional distribution given by the generator is encouraged to be more continuous from the perspective of K-Lipschitz continuous conditional distributions.

\section{Experiments}
In this section, we empirically evaluate the proposed regularization term on two synthetic experiments and one image generation experiment. Additional experiment results and details can be found in the supplementary matrials. For fair comparison, unless otherwise specified, we always try to use the same network architectures, evaluation metrics, and hyper-parameters as CcGAN for GR-cGAN and other baseline models based on CcGAN's open-source implementation\footnote{\href{https://github.com/UBCDingXin/improved\_CcGAN}{https://github.com/UBCDingXin/improved\_CcGAN.}}.We also publish our code with 
GitHub\footnote{\href{https://github.com/gpcgan/GR-cGAN}{https://github.com/gpcgan/GR-cGAN.}}.

\subsection{Circular 2-D Gaussians}
\label{sec:cir_2d}
We test generator regularization on the synthetic data generated from 2-D Gaussians with different means, and compare our results with CcGAN \citep{ding2021ccgan}.

\subsubsection{Experimental Setup}
We generate a synthetic dataset using the same method as presented in CcGAN to show the effect of generator regularization. The data is generated from 2-D Gaussians with different means. The condition $\boldsymbol{x}$ has a dimension of one which measures the polar angle of a given data point and the dependency $\boldsymbol{y}$ has a dimension of two. Given $\boldsymbol{x}\in [0,2\pi]$, we construct $\boldsymbol{y}$ such that it follows a 2-D Gaussian distribution, specifically,
$$
\boldsymbol{y}\sim \mathcal{N}(\boldsymbol{\mu}_{\boldsymbol{x}},\boldsymbol{\Sigma}) \text{ with }\boldsymbol{\mu}_{\boldsymbol{x}}=\left(\begin{array}{c}
R\cdot\sin(\boldsymbol{x}) \\
R\cdot\cos(\boldsymbol{x})
\end{array}\right)
$$
and
$$
\boldsymbol{\Sigma}=\tilde{\sigma}^{2} I_{2 \times 2}=\left(\begin{array}{cc}
\tilde{\sigma}^{2} & 0 \\
0 & \tilde{\sigma}^{2}
\end{array}\right).
$$

The distribution of $ \boldsymbol{y} $ is a two-dimensional Gaussian distribution, with its center located on a circle with a radius of $ R $, and the position of the center on the circle is controlled by $\boldsymbol {x}$. 

For a thorough analysis, we study several different settings for $\boldsymbol{x}$ when generating the dataset. In Section \ref{sec:full_dataset}, $\boldsymbol{x}$ is evenly distributed in the range of $[0, 2\pi]$. In Section \ref{sec:partial_dataset}, we choose a subset of $[0,2\pi]$ for training and evaluate how well the models can generalize to the gaps that are absent during training.

\subsubsection{Full Dataset}
\label{sec:full_dataset}
\begin{figure}[H]
\centering 
\subfigure[]
{
\begin{minipage}{0.3\columnwidth}
\centering                                           
\includegraphics[height=\columnwidth]{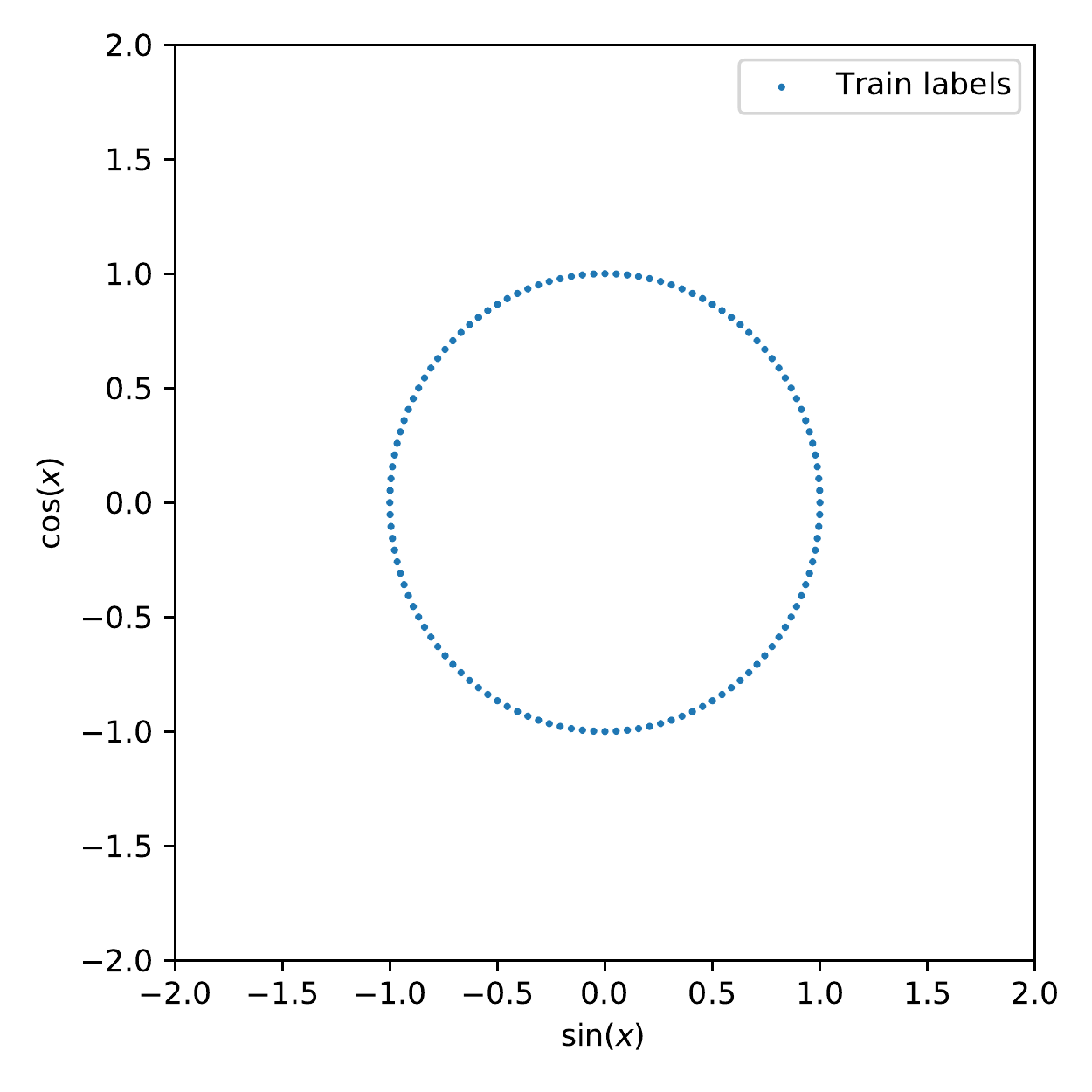}   
\label{fig:full_train_angles}
\end{minipage}
}
\subfigure[]
{
\begin{minipage}{0.3\columnwidth}
\centering                                           
\includegraphics[height=\columnwidth]{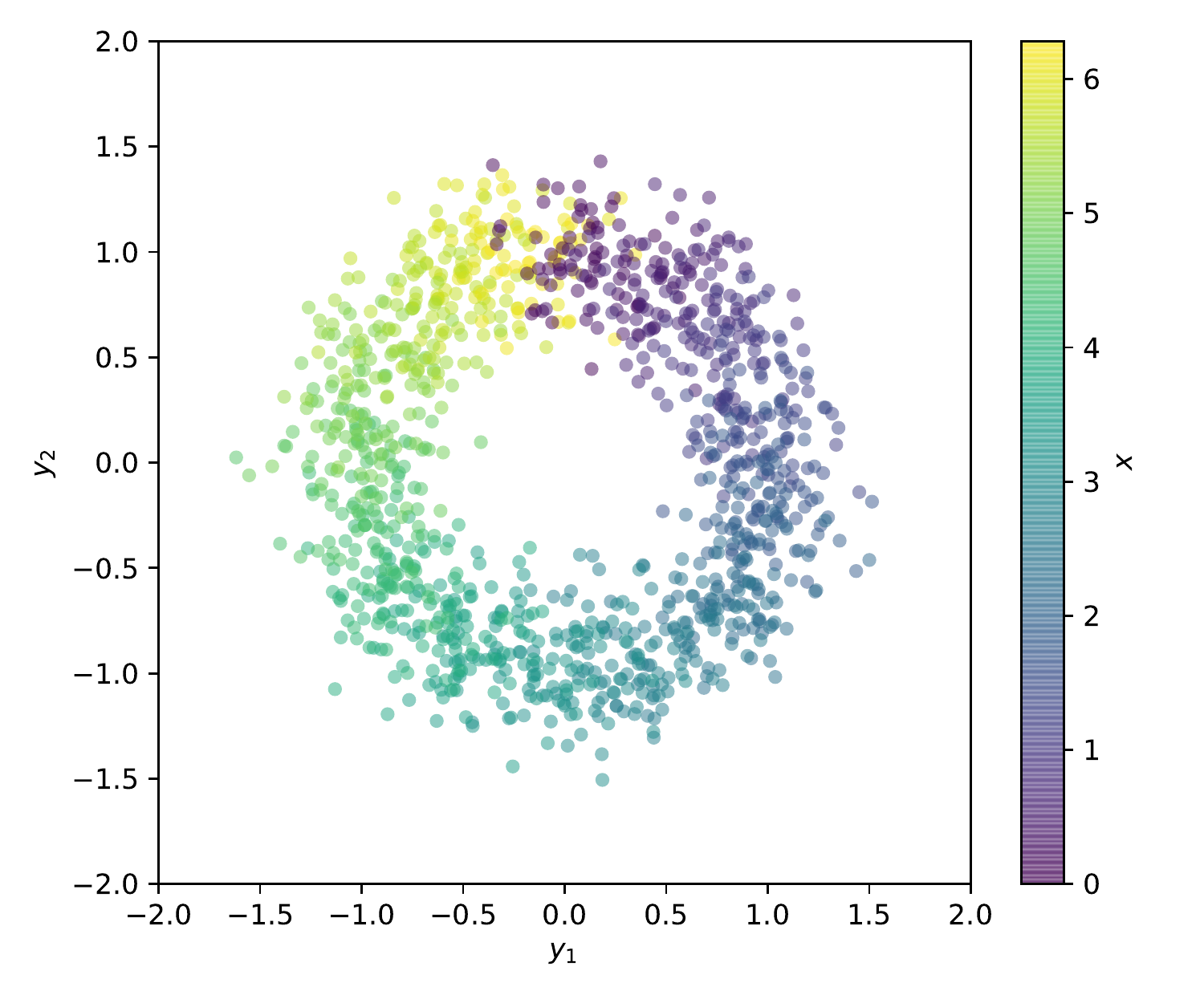}    
\label{fig:full_trainingset}
\end{minipage}}
\caption{(a) plots the locations of the means of the 120 Gaussians. (b) illustrates 1,200 randomly chosen samples from the training set.}
\label{fig:full}
\end{figure}

The positions of the train labels are shown in Figure \ref{fig:full_train_angles}. To generate a training set, for each x in the train labels, 10 samples are generated. Figure \ref{fig:full_trainingset} shows 1,200 training samples. We set $R=1$ and $\tilde{\sigma}^2=0.2$. We use the CcGAN (HVDL) and CcGAN (SVDL) models in CcGAN as baseline models. We also consider the degenerated case of the proposed GR-cGAN (Degenerated GR-cGAN) by setting the generator regularization coefficient $\lambda$ to zero. For the GR-cGAN model, we use the loss term given in Equation \ref{eq: gp_original} where $\lambda = 0.02$. All these models are trained the same dataset for 6,000 iterations. See supplementary materials for details. We end with a discussion of the implications from these experiments.

\textbf{Evaluation metrics and quantitative results:} With the trained models, we use the same steps and evaluation metrics as in CcGAN. We choose 360 values of $\boldsymbol{x}$ evenly from the interval $[0,2\pi]$. For each model, given a value of $\boldsymbol{x}$, we generate 100 samples, yielding 36,000 fake samples in total. We evaluate the quality of these fake samples.

A circle with $(\sin(\boldsymbol{x}), \cos(\boldsymbol{x}))$ as the center and $2.15\tilde{\sigma}$ as the radius can enclose about 90\% of the volume inside the pdf of $\mathcal{N}(\boldsymbol{\mu}_{\boldsymbol{x}},\boldsymbol{\Sigma})$. We define a fake sample $\boldsymbol{y}$ as a high quality sample if its Euclidean distance from $\boldsymbol{y}$ to $(\sin(\boldsymbol{x}), \cos(\boldsymbol{x}))$ is smaller than $2.15\tilde{\sigma}=0.43$. A mode (i.e., a Gaussian) is recovered if at least one high quality sample is generated. For the conditional distribution given by the generator, (i.e., the distribution of $G(\boldsymbol{x},\boldsymbol{z})$ with $\boldsymbol{z}\sim p_z(\boldsymbol{z})$), we assume this distribution is Gaussian and estimate its mean and covariance using 100 fake samples, denoted by $\boldsymbol{\mu}_{\boldsymbol{x}}^{G}$ and $\boldsymbol{\Sigma}_{\boldsymbol{x}}^{G}$ respectively. We compute the \textbf{2-Wasserstein Distance (W2)} \cite{peyre2019computational} between the true conditional distribution and the distribution given by the generator, in other words, the 2-Wasserstein Distance between
$$\mathcal{N}\left(\left(\begin{array}{c}
R\cdot\sin(\boldsymbol{x}) \\
R\cdot\cos(\boldsymbol{x})
\end{array}\right),\tilde{\sigma}^{2} I_{2 \times 2}\right)\text{ and }\mathcal{N}(\boldsymbol{\mu}_{\boldsymbol{x}}^{G},\boldsymbol{\Sigma}_{\boldsymbol{x}}^{G}).
$$
The whole experiment is repeated three times and the averaged values of the metrics are reported in Table \ref{tab:full_table} over three repetitions. We see that GR-cGAN demonstrates competitive performances against CcGAN, especially in terms of the 2-Wasserstein distance.

\begin{table*}[ht!] 
\begin{center}
\begin{small}
\begin{sc}
\begin{tabular}{lcccr}
\toprule
Model & \% High Quality & \% Recovered Mode & 2-Wasserstein Dist. \\
\midrule
CcGAN (HVDL)            & \textbf{95.9}      & 100      & $3.79\times 10^{-2}$ \\
CcGAN (SVDL)            & 91.8      & 100      & $5.37\times 10^{-2}$\\
Deg. GR-cGAN            & \textbf{95.9}      & 100      & $3.79\times 10^{-2}$ \\
GR-cGAN                 & 93.7      & \textbf{100}      & $\boldsymbol{2.63\times 10^{-2}}$ \\
\bottomrule
\end{tabular}
\end{sc}
\end{small}
\caption{Evaluation metrics for the full dataset experiments.\label{tab:full_table} \textmd{The metrics of 36,000 fake samples generated from each model over three repetitions are given. Larger values of ``\% Recovered Mode" are better, while smaller values of ``2-Wasserstein Dist." are preferred. Note that the larger values of ``\% High Quality." does not completely mean that the GAN model is better, because the samples generated by a GAN whose distribution is concentrated to a point located within the threshold will also be considered as high-quality.}}
\end{center}
\end{table*}

\begin{figure}[H]
\centering 
\subfigure[CcGAN (HVDL)]
{
\begin{minipage}{0.3\columnwidth}
\centering                                           
\includegraphics[height=\columnwidth]{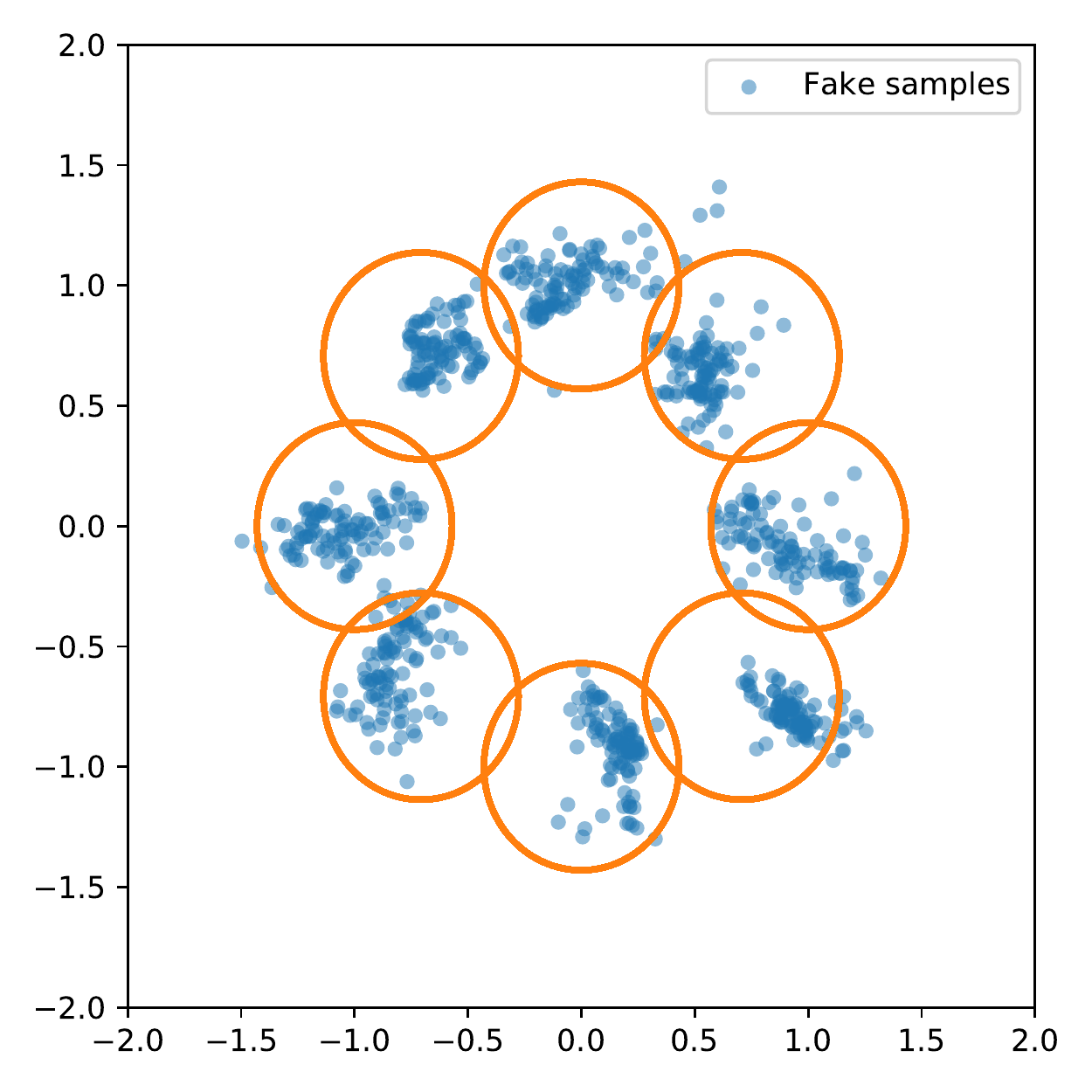}
\end{minipage}
}
\subfigure[CcGAN (SVDL)]
{
\begin{minipage}{0.3\columnwidth}
\centering                                           
\includegraphics[height=\columnwidth]{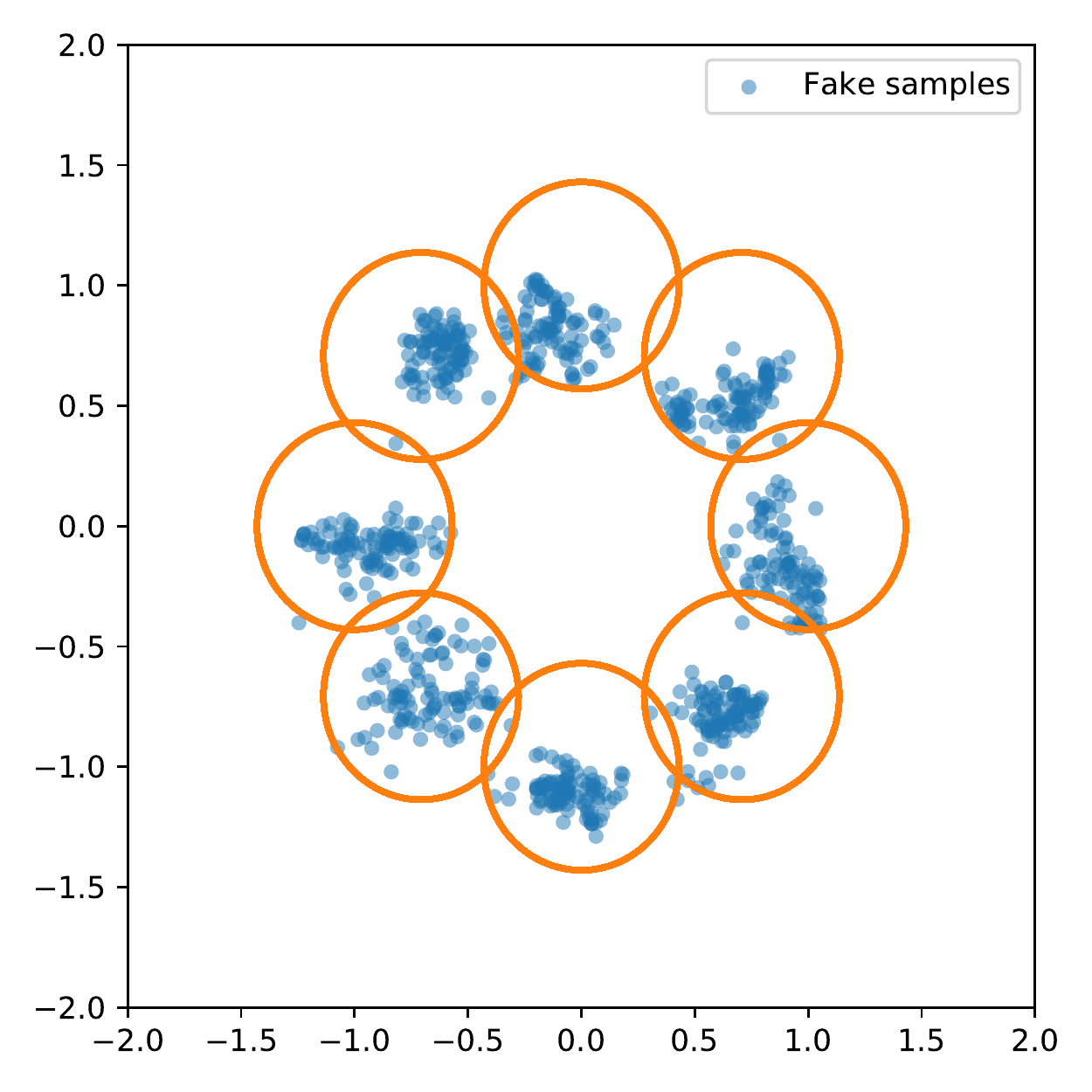}
\end{minipage}}

\subfigure[Degenerated GR-cGAN]
{
\begin{minipage}{0.3\columnwidth}
\centering                                           
\includegraphics[height=\columnwidth]{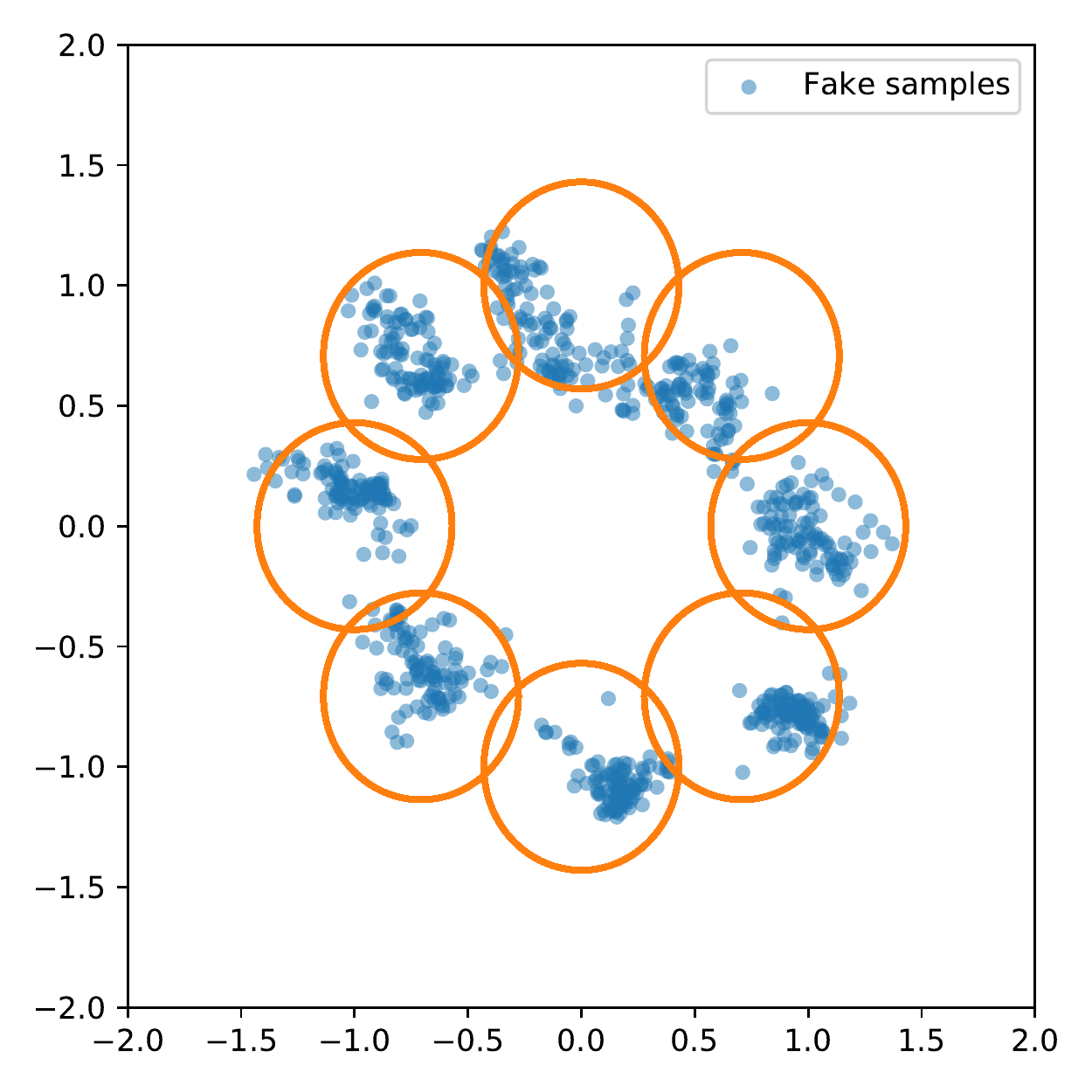}
\end{minipage}
}
\subfigure[GR-cGAN]
{
\begin{minipage}{0.3\columnwidth}
\centering                                           
\includegraphics[height=\columnwidth]{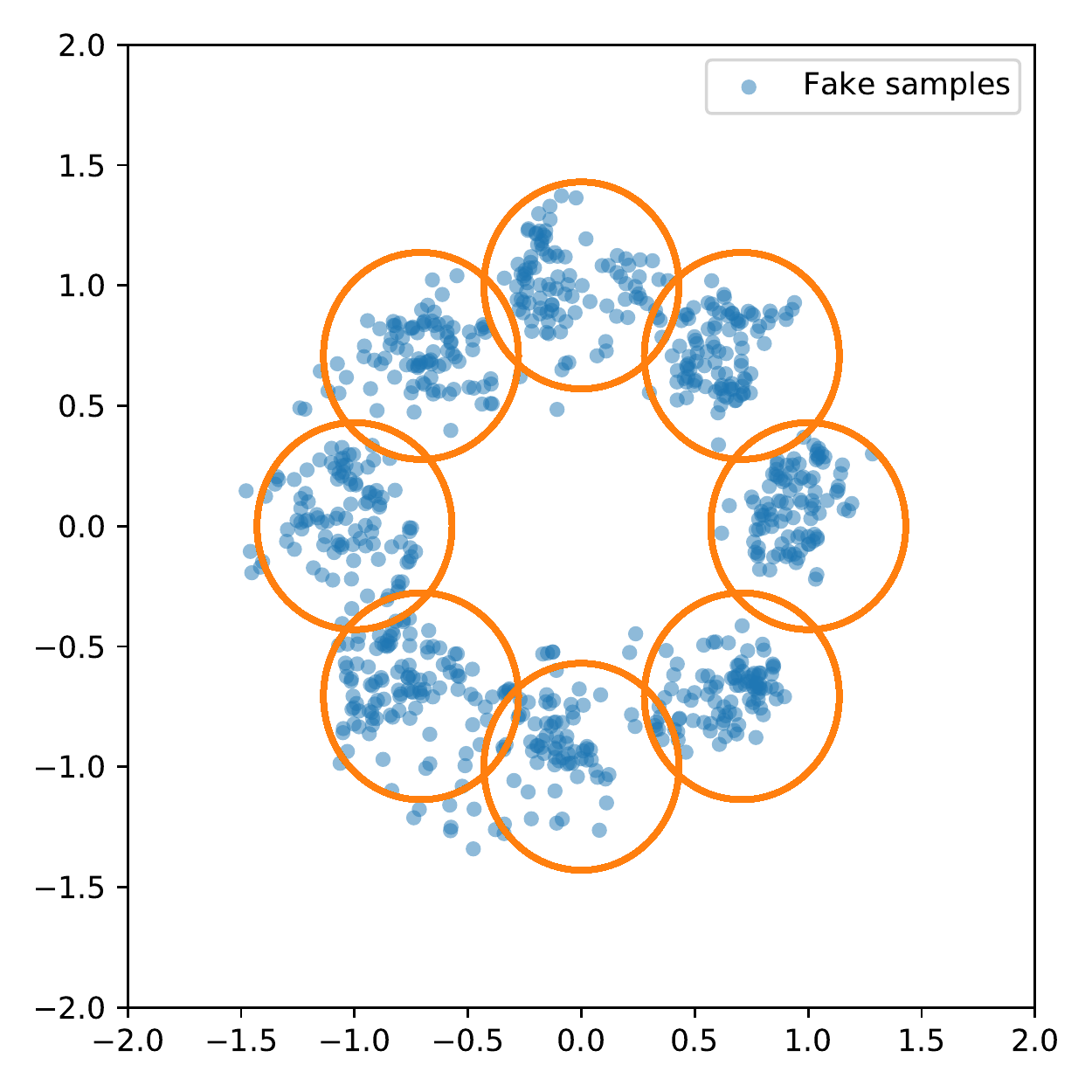}
\end{minipage}}
\caption{Visual results of the Circular 2-D Gaussians experiments on the full dataset. \textmd{For each subfigure, we generate 100 fake samples using each model at each of the 8 means that are absent from the training set. The blue dots represent the fake samples. For each mean $\boldsymbol{x}$ given, the circle locates at $(\sin(\boldsymbol{x}), \cos(\boldsymbol{x}))$ and has a radius of $2.15\tilde{\sigma}$, which can cover about 90\% of the volume inside the pdf of $\mathcal{N}(\boldsymbol{\mu}_{\boldsymbol{x}},\boldsymbol{\Sigma})$.}\label{fig:result_full}}
\label{fig:full}
\end{figure}

\textbf{Visual results:} We select 8 angles that do not exist in the training set. For each angles $\boldsymbol{x}$ selected,  we use all the models to generate 100 fake samples. Furthermore, we plot the circle with $(\sin(\boldsymbol{x}), \cos(\boldsymbol{x}))$ as the center and $ 2.15\tilde{\sigma}$ as the radius to indicate the true conditional distribution $\mathcal{N}(\boldsymbol{\mu}_{\boldsymbol{x}},\boldsymbol{\Sigma})$. The results are given in Figure \ref{fig:result_full}. Fake samples from our method better match the true samples when compared to the other methods.

\subsubsection{Partial Dataset}
\label{sec:partial_dataset}
\begin{figure}[ht!]
\centering 
\subfigure[]
{
\begin{minipage}{0.3\columnwidth}
\centering                                           
\includegraphics[height=\columnwidth]{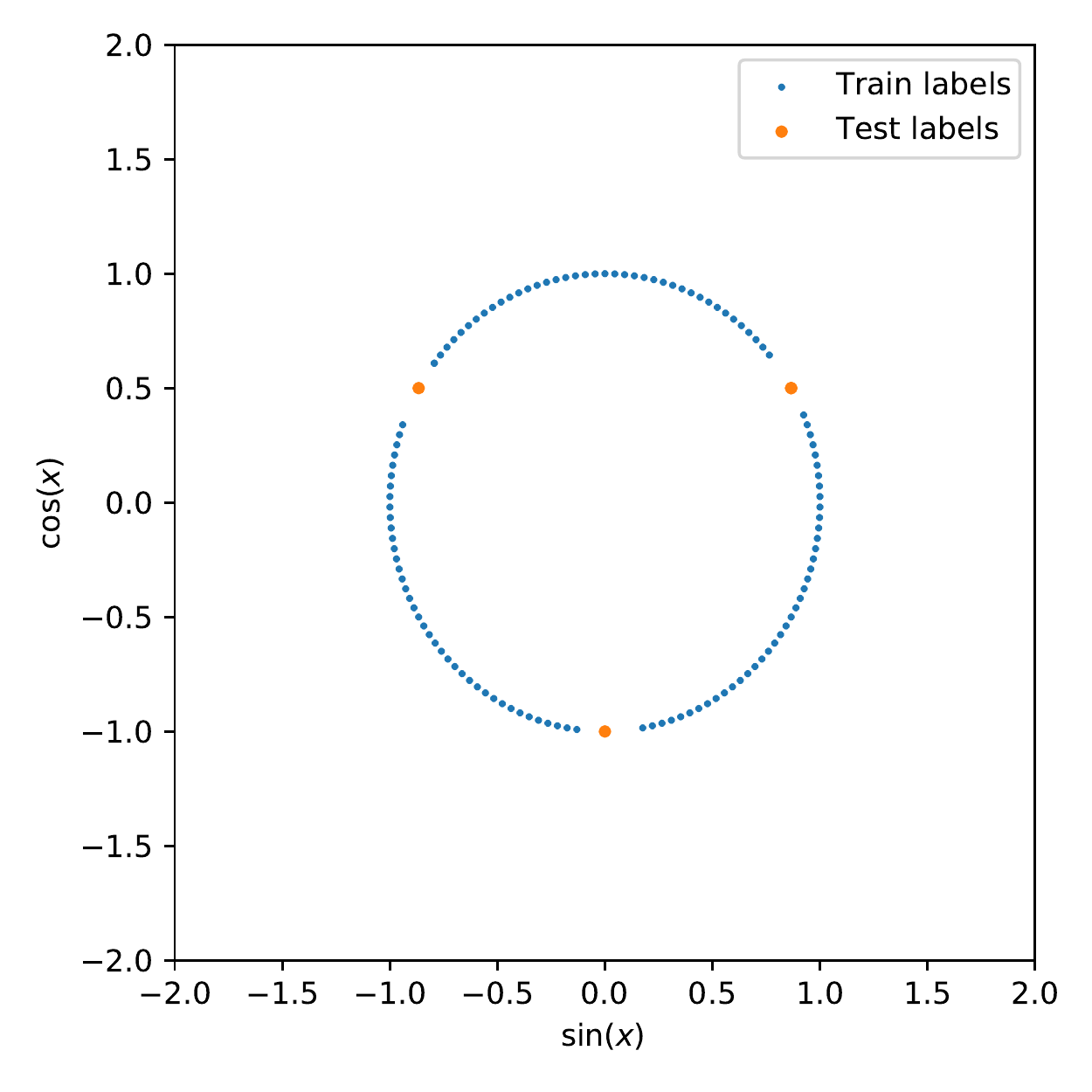}   
\label{fig:part_label}
\end{minipage}
}
\subfigure[]
{
\begin{minipage}{0.3\columnwidth}
\centering                                           
\includegraphics[height=\columnwidth]{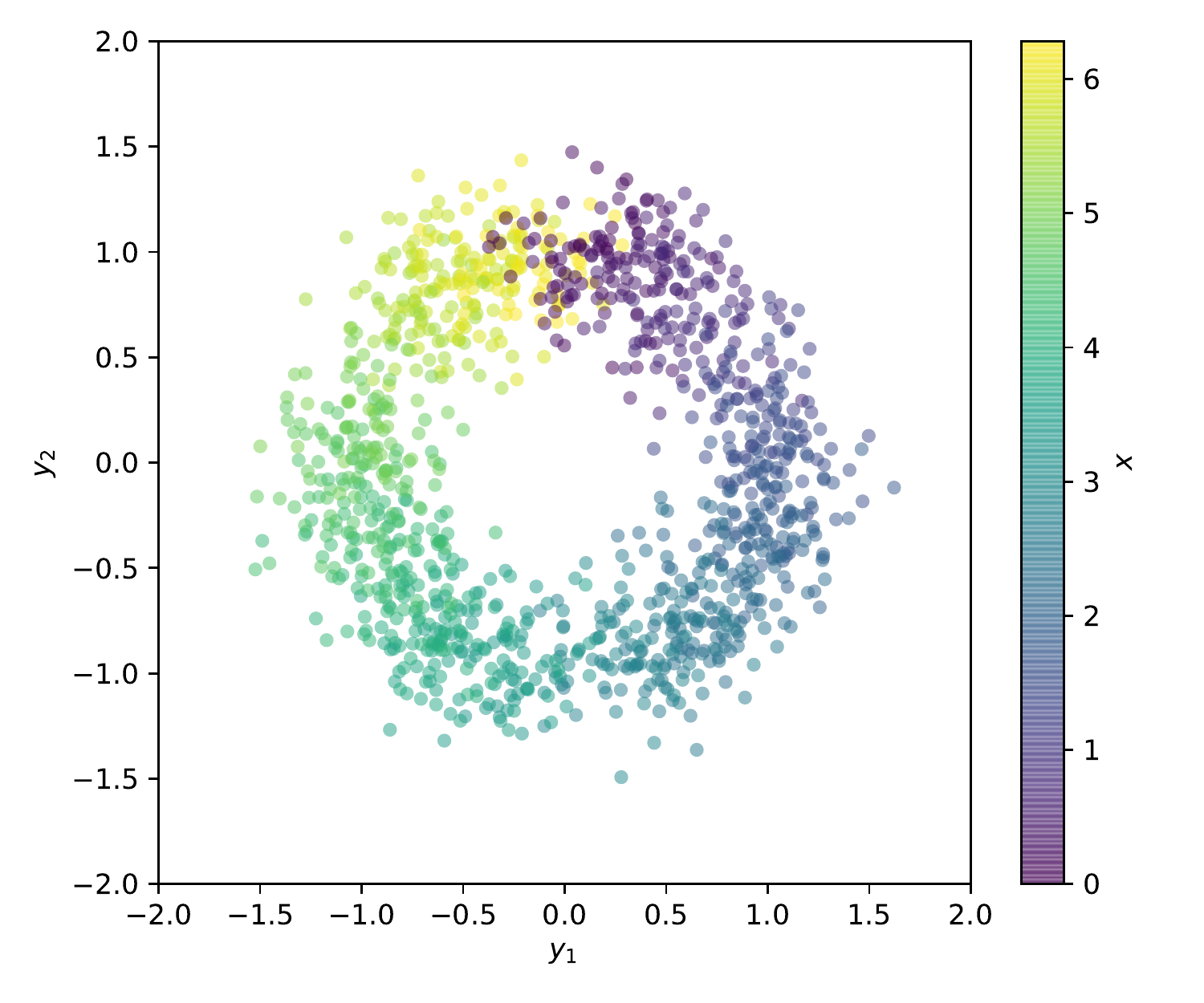}    
\label{fig:part_trainingset}
\end{minipage}}
\caption{(a) illustrates the train labels and test labels. \textmd{Given a label $\boldsymbol{x}$, we plot a dot at $(\sin(\boldsymbol{x},\cos(\boldsymbol{x}))$. The blue dots correspond to the train labels, while the orange dots correspond to the test labels.} (b) gives the 1,200 samples in the training set. \textmd{The color of each dot represents which train labels it belongs to.}}
\label{fig:part}
\end{figure}

To examine the robustness of each model to the presence of gaps in the training set, we intentionally select a subset of $[0, 2\pi]$ and only train the models on the subset. Specifically, we set three gaps with a length of $\pi/12$, and remove these gaps from the range $[0,2\pi]$ to get a subset of $[0,2\pi]$. These three gaps are non-overlapping and are evenly located in $[0,2\pi]$. We set $\boldsymbol{x}$ to 120 different values that are evenly arranged in the subset, which are then used as the train labels. Each value of $\boldsymbol{x}$ is the mean of a Gaussian distribution. For each gap, we use the angle in the middle of the gap as the test label to evaluate the performance of the models. Thus, the three gaps correspond to three test labels. The positions of the train labels and test labels are shown in Figure \ref{fig:part_label}. Please refer to the Supplementary Materials or the released code for the specifics of retrieving these labels. To generate a training set, for each $\boldsymbol{x}$ in the train labels, 10 samples are generated. We denote this training set as the partial dataset. For $R$ and $\tilde{\sigma}^2$, we used the same value as in Section \ref{sec:full_dataset}, i.e. $R=1$ and $\tilde{\sigma}^2=0.2$. Figure \ref{fig:part_trainingset} shows 1,200 training samples on the partial dataset. The network structure and training parameters are consistent with those in Section \ref{sec:full_dataset}.

\textbf{Results:} We used the same evaluation metrics as in Section \ref{sec:full_dataset}. We generate 100 fake samples on each test label and calculate the value of the metrics. The results are given in Table \ref{tab:part_table}.

\begin{table*}[h] 
\begin{center}
\begin{small}
\begin{sc}
\begin{tabular}{lcccr}
\toprule
Model & \% High Quality & \% Recovered Mode & 2-Wasserstein Dist. \\
\midrule
CcGAN (HVDL)            & 91.0      & 100      & $3.77\times 10^{-2}$ \\
CcGAN (SVDL)            & \textbf{95.7}   & 100      & $3.59\times 10^{-2}$\\
Deg. GR-cGAN            & 93.9      & 100      & $4.51\times 10^{-2}$ \\
GR-cGAN                 & 93.5      & \textbf{100}      & $\boldsymbol{3.06\times 10^{-2}}$ \\
\bottomrule
\end{tabular}
\end{sc}
\end{small}
\caption{Evaluation metrics for the partial dataset experiments. \label{tab:part_table}\textmd{The metrics of 36,000 fake samples generated from each model over three repetitions are given. The metrics of 36,000 fake samples generated from each model over three repetitions are given. Larger values of ``\% Recovered Mode" are better, while smaller values of ``2-Wasserstein Dist." are preferred. Note that the larger values of ``\% High Quality." does not completely mean that the GAN model is better, because the samples generated by a GAN whose distribution is concentrated to a point located within the threshold will also be considered as high-quality.}}
\end{center}
\end{table*}

We plot these fake samples in Figure \ref{fig:part}. For each test label $\boldsymbol{x}$, a circle that covers about 90\% of the volume inside the pdf of $\mathcal{N}(\boldsymbol{\mu}_{\boldsymbol{x}},\boldsymbol{\Sigma})$ is also plotted.

\begin{figure}[ht!]
\centering 
\subfigure[CcGAN (HVDL)]
{
\begin{minipage}{0.3\columnwidth}
\centering                                           
\includegraphics[height=\columnwidth]{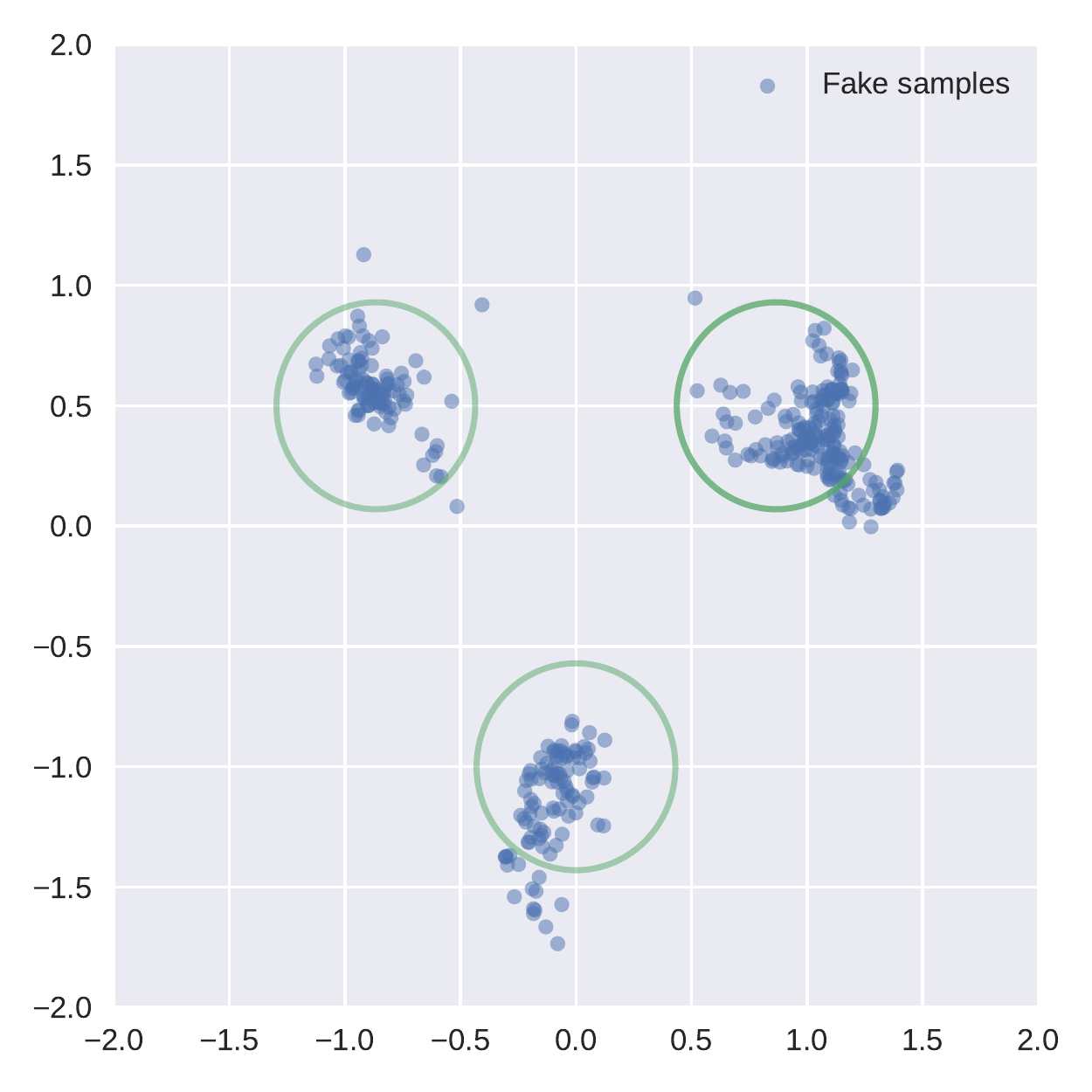}   
\end{minipage}
}
\subfigure[CcGAN (SVDL)]
{
\begin{minipage}{0.3\columnwidth}
\centering                                           
\includegraphics[height=\columnwidth]{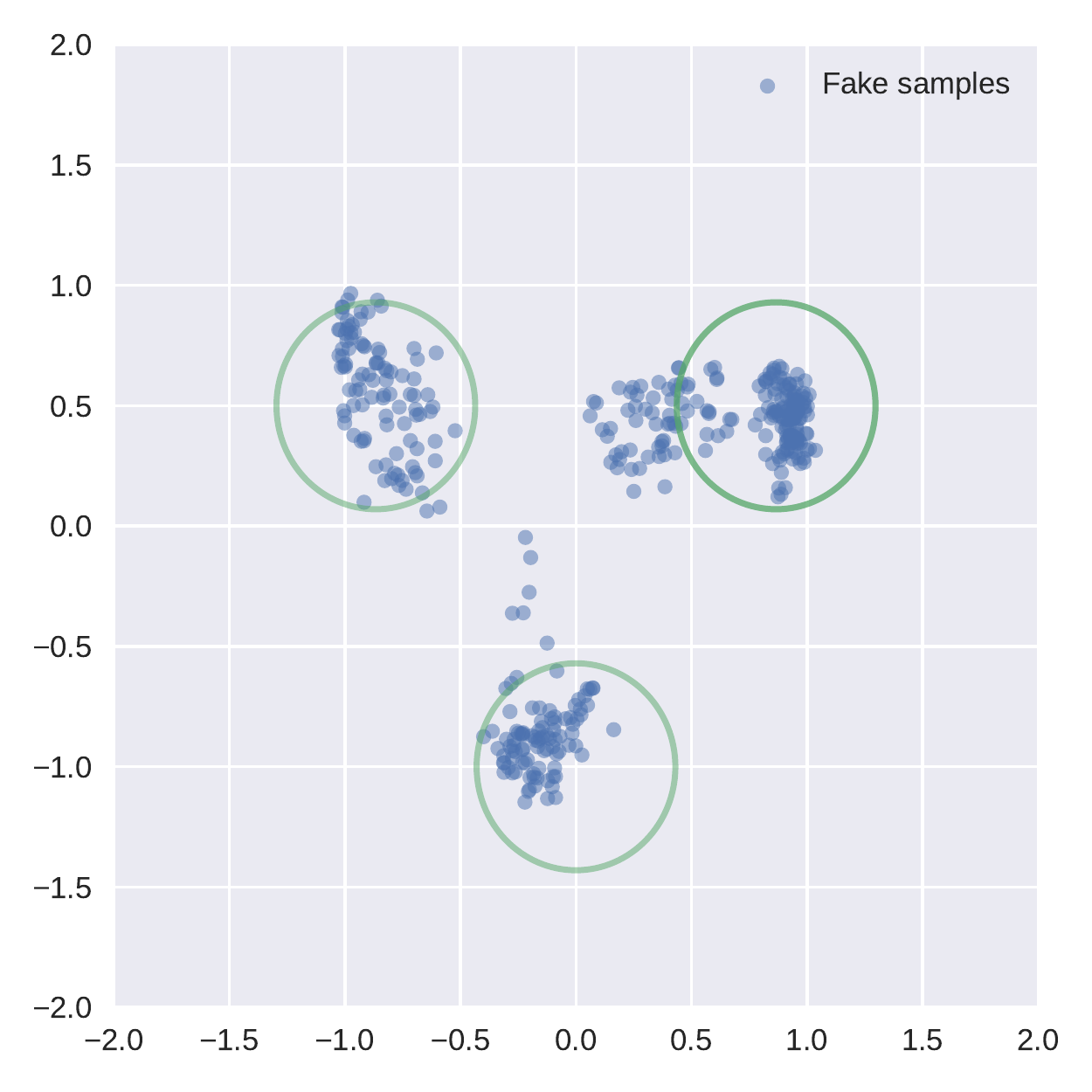}
\end{minipage}}

\subfigure[Degenerated GR-cGAN]
{
\begin{minipage}{0.3\columnwidth}
\centering                                           
\includegraphics[height=\columnwidth]{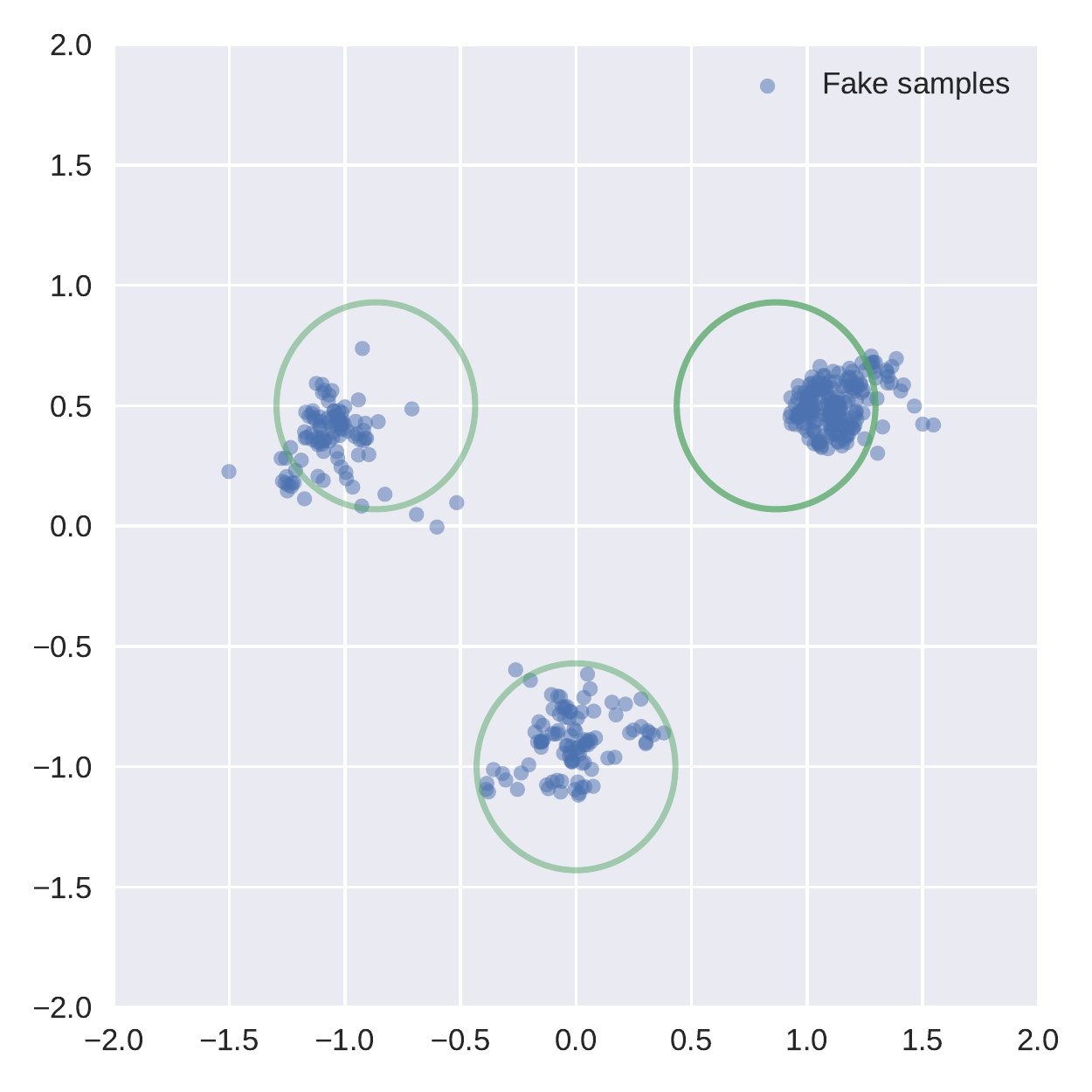}
\end{minipage}
}
\subfigure[GR-cGAN]
{
\begin{minipage}{0.3\columnwidth}
\centering                                           
\includegraphics[height=\columnwidth]{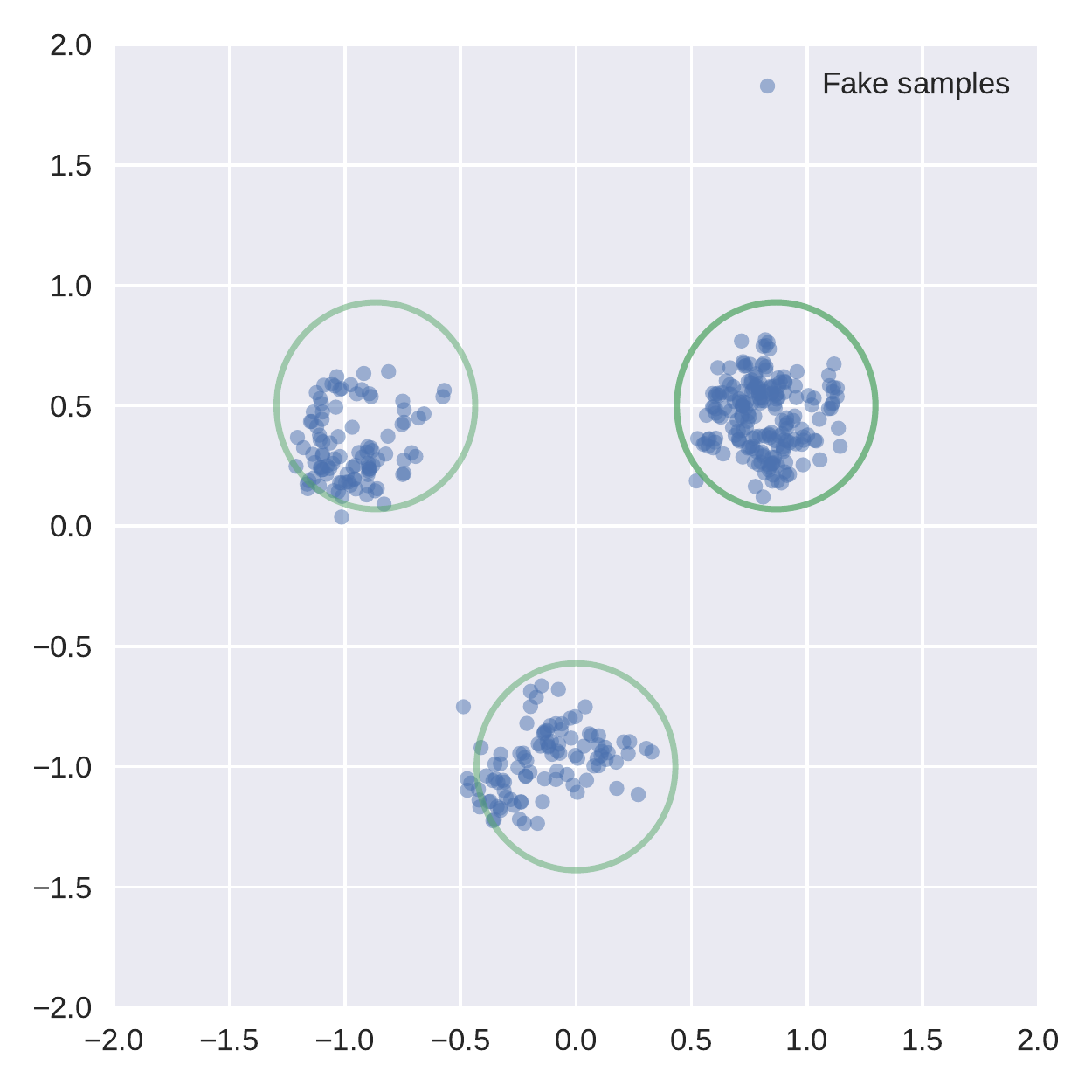}
\end{minipage}}
\caption{Visual results of the Circular 2-D Gaussians experiments on partial dataset. \textmd{For each subfigure, we generate 100 fake samples using each GAN model at each of 3 labels in the test labels. The blue dots represent the fake samples. For each mean $\boldsymbol{x}$ in the test labels, a circle that can cover about 90\% of the volume inside the pdf of $\mathcal{N}(\boldsymbol{\mu}_{\boldsymbol{x}},\boldsymbol{\Sigma})$ is plotted.}}
\label{fig:part}
\end{figure}

GR-cGAN achieves visually reasonable results, and performs good from the perspective of the evaluation metrics. It is a good property that the generator can give reasonable fake samples even when given a label on a gap. GR-cGAN can be used in the case where there are missing labels in the training set. For example, in the task of generating photos of people with a given character description, if we only have samples of ``young and happy" and ``old and sad", we can use GR-cGAN to generate ``old and happy" images.

\subsection{Multivariate Gaussian}
\label{sec:mul_gaussian}
We demonstrate the performance of generator regularization to learn and generate a multivariate gaussian distribution.

\subsubsection{Experimental Setup}

The distribution we use is a $k$-dimensional multivariate Gaussian distribution given by $\mathbf{X} \sim \mathcal{N}(\boldsymbol{\mu}, \mathbf{\Sigma})$, where $\boldsymbol{\mu}$ is the mean vector and $\boldsymbol{\Sigma}$ is the covariance matrix. Denote the first $p$ dimensions of $\mathbf{X}$ as $\boldsymbol{x}$ and the last $k-p$ dimensions of $\textbf{X}$ as $\boldsymbol{y}$. The benifit of using such distribution is that we can effectively evaluate the true conditional distribution $p_r(\boldsymbol{y}|\boldsymbol{x})$.

In the experiment we set $k=10$ and $p=8$, with the dimension of $\boldsymbol{y}$ is $k-p=2$. The parameters of the distribution, $\boldsymbol{\mu}$ and $\boldsymbol{\Sigma}$, are pre-specified. We sample $N=1,000$ iid copies of $\mathbf{X}$, denoted by $\{\boldsymbol{X}_i\}_{i=1}^N$. The training set is $\{(\boldsymbol{x}_i, \boldsymbol{y}_i)\}_{i=1}^N$, where $\boldsymbol{x}_i$ denotes the first $p$ dimensions of $\boldsymbol{X}_i$, and $\boldsymbol{y}_i$ is the last $k-p$ dimensions of $\boldsymbol{X}_i$. We use vanilla cGAN as a baseline model. The same dataset is used for all the experiment repetitions. For a fairness comparison, we use a same net architecture for all the models. The details are given in Supplementary Materials.

\subsubsection{Results}
\label{sec:multi_gaussian_result}
\begin{figure}[ht!]
\centering 
\subfigure[Discriminator view of cGAN]
{
\begin{minipage}{0.4\columnwidth}
\centering                                           
\includegraphics[height=0.525\columnwidth]{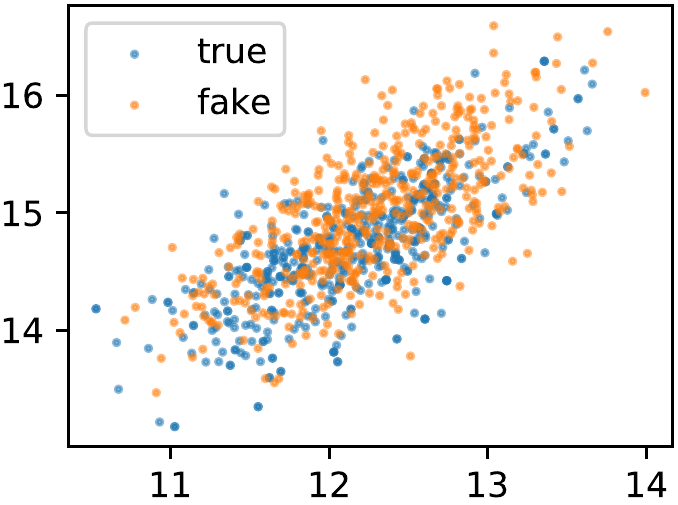}
\label{fig:highdim_cglobal}
\end{minipage}
}
\subfigure[Discriminator view of GR-cGAN]
{
\begin{minipage}{0.4\columnwidth}
\centering                                           
\includegraphics[height=0.525\columnwidth]{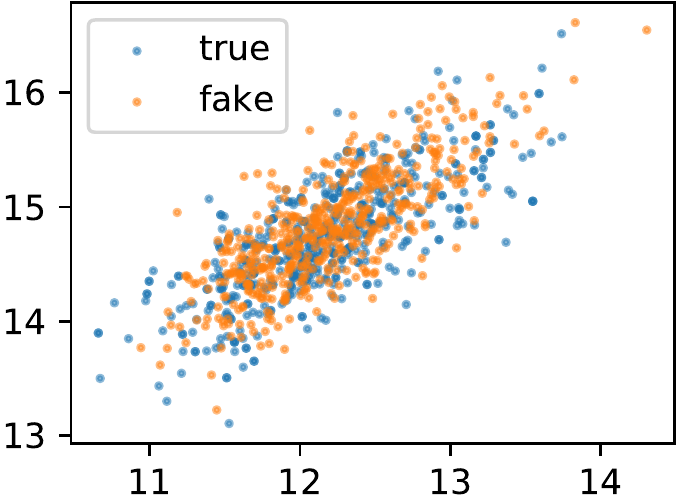}
\label{fig:highdim_gpglobal}
\end{minipage}}

\subfigure[Conditional distribution of cGAN]
{
\begin{minipage}{0.4\columnwidth}
\centering                                           
\includegraphics[height=0.525\columnwidth]{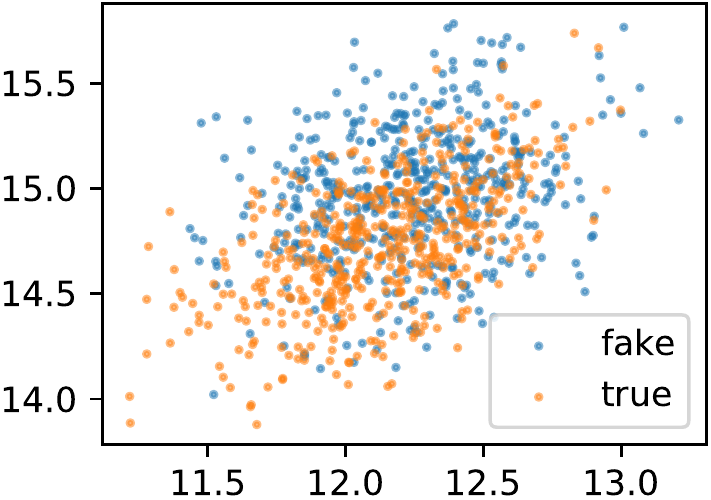}
\label{fig:highdim_clocal}
\end{minipage}
}
\subfigure[Conditional distribution of GR-cGAN]
{
\begin{minipage}{0.4\columnwidth}
\centering                                           
\includegraphics[height=0.525\columnwidth]{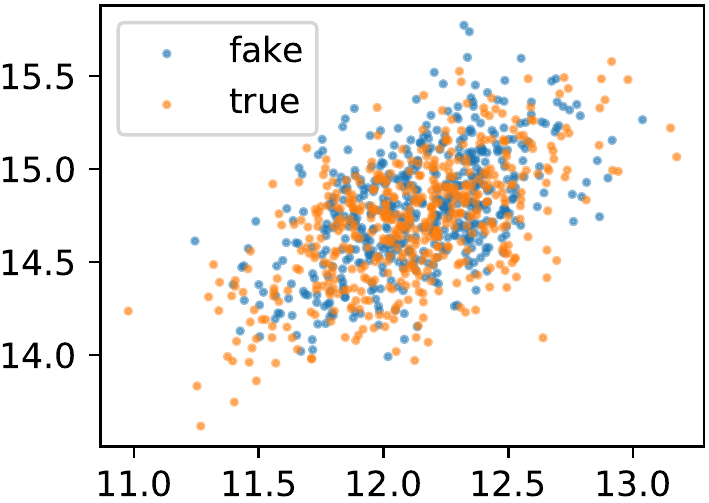}
\label{fig:highdim_gplocal}
\end{minipage}}
\caption{Visual results of the multivariate Gaussian experiment. \textmd{In (a) and (b), from the discriminator's view, there's only little difference between the real samples and fake samples. But in (c), the conditional distribution given by the generator deviates from the true conditional distribution. In (d), by using generator regularization, the conditional distribution given by the generator gets closer to the real conditional distribution.}}
\label{fig:full}
\end{figure}

We first show what the discriminator sees during the training process. We use the same steps to get a batch of true samples and fake samples as we train a cGAN. We sample a batch of training samples ($\boldsymbol{x}_i, \boldsymbol{y}_i$)'s from the training set. The term $\boldsymbol{y}_i$'s are two dimensional. We can plot the location of $\boldsymbol{y}_i$'s in Figure \ref{fig:highdim_cglobal} and \ref{fig:highdim_gpglobal}, with these points marked as true samples. For each $\boldsymbol{x}_i$, we sample one noise $\boldsymbol{z}_i$ from the noise distribution $p_z(\boldsymbol{z})$. We plot the points $\left(G_1(\boldsymbol{x}_i, \boldsymbol{z}_i), G_2(\boldsymbol{x}_i, \boldsymbol{z}_i)\right)$'s in Figure \ref{fig:highdim_cglobal} and \ref{fig:highdim_gpglobal}, where $G_i(\boldsymbol{x}_i, \boldsymbol{z}_i)$ is the $i$-th dimension of $G_1(\boldsymbol{x}_i, \boldsymbol{z}_i)$. These points are named fake samples. There's only subtle difference between the distribution of true samples and fake samples for both GAN models.

Then we compare the distribution given by the generator with the true conditional distribution $p_r(\boldsymbol{y}|\boldsymbol{x})$. Denote the first $p$ dimensions of $\boldsymbol{\mu}$ as $\boldsymbol{\mu}_{1:p}$. We set the label that we take condition on as $\boldsymbol{x}=\boldsymbol{\mu}_{1:p}$. (The comparison on more conditions are further given in Supplementary Materials.) Using the generator, we generate 250 fake samples $G(\boldsymbol{\mu}_{1:p}, \boldsymbol{z}_i)$, for $i=1,2,\ldots, 250$, with $\boldsymbol{z}_i$ sampled from $p_{\boldsymbol{z}}(\boldsymbol{z})$. We also get 250 samples from the true conditional distribution $p_r(\boldsymbol{y}|\boldsymbol{x}=\boldsymbol{\mu}_{1:p})$ and denote these samples as true samples. We compare the distribution of fake samples with true samples in Figure \ref{fig:highdim_clocal} and \ref{fig:highdim_gplocal}.

\subsection{RC-49}
We further evaluate our model on the RC-49 dataset, which consists of $44,051$ $64 \times 64$ rendered RGB images of $49$ 3-D chair models at different yaw angles. For this task, the generative conditions are $899$ yaw angles ranging from $0.1$ degrees to $89.9$ degrees with a step size of $0.1$. For fair comparison, we adapt the label embedding module in CcGAN for both the vanilla cGAN baseline and GR-CGAN. Note that the original CcGAN paper selects a yaw angle for training if its last digit is odd. In other words, the gaps between adjacent training conditions are $0.2$. To compare the capabilities of models under more challenging settings, we increase the gap to $20$ and report the results in Table \ref{tab:rc49}. We also include a version of GR-cGAN where we only penalize the gradients at the training samples instead of along the interpolations as an additional baseline for ablation studies.

\begin{table*}[ht!]
\label{tab: rc49_gap20}
\begin{center}
\begin{small}
\begin{sc}
\begin{tabular}{lcccr}
\toprule
Model & Intra-FID $\downarrow$ & Label Score $\downarrow$ & Diversity $\uparrow$ \\
\midrule
cGAN            & 0.4179 $\pm$ 0.0907      & 2.5197 $\pm$ 0.8249     & 2.7724 $\pm$ 0.1209 \\
CcGAN (SVDL)            & 0.4391 $\pm$ 0.1149     & 4.1077 $\pm$ 1.9512     & \textbf{2.7772 $\pm$ 0.1357} \\
GR-cGAN (no interpolation)            & 0.4215 $\pm$ 0.0812      & 1.9661 $\pm$ 0.6873    & 2.7212 $\pm$ 0.0548 \\
GR-cGAN                 & \textbf{0.3982 $\pm$ 0.1020}     & \textbf{1.6628 $\pm$ 0.7594}      & 2.7581 $\pm$ 0.0997 \\
\bottomrule
\end{tabular}
\caption{Performance comparisons on the RC-49 dataset when gap is set to 20. \label{tab:rc49}\textmd{$\uparrow$ indicates higher values are preferred, while $\downarrow$ indicates lower values are preferred.}}
\end{sc}
\end{small}
\end{center}
\end{table*}

\begin{figure*}[h]
\centering 
\subfigure[Vanilla cGAN]
{
\begin{minipage}{0.4 \columnwidth}
\centering                                           
\includegraphics[height=\columnwidth]{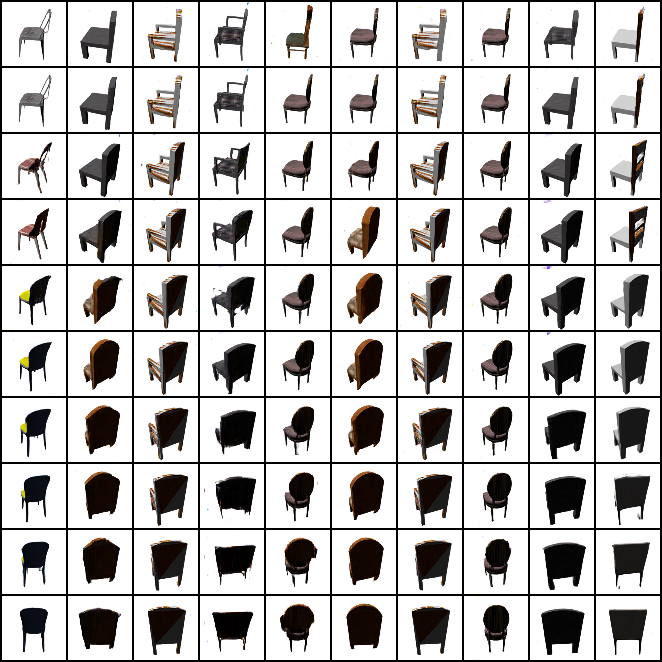} 
\label{fig: rc49_gap20_cgan}
\end{minipage}
}
\subfigure[GR-cGAN]
{
\begin{minipage}{0.4 \columnwidth}
\centering                                           
\includegraphics[height=\columnwidth]{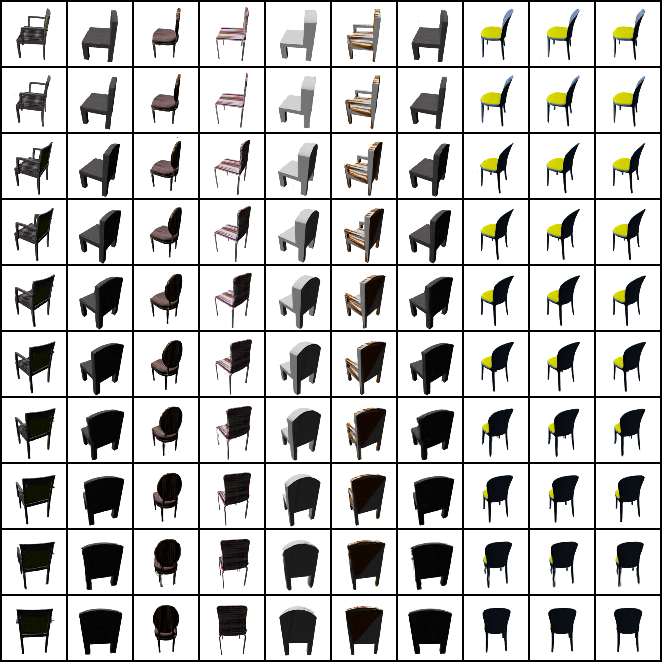}  
\label{fig: rc49_gap20_cgangp}
\end{minipage}
}
\subfigure[CcGAN]
{
\begin{minipage}{0.4 \columnwidth}
\centering                                           
\includegraphics[height=\columnwidth]{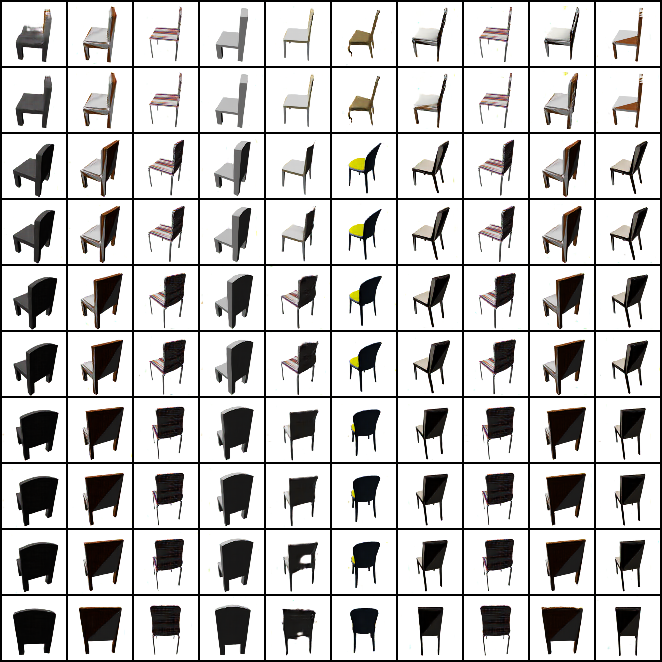}   
\label{fig: rc49_gap20_ccgan}
\end{minipage}}
\caption{Comparisons of the three models on the RC-49 dataset when the gap is set to $20$. \textmd{The rows corresponds to different yaw angles, while each column uses the same latent noise $z$. The proposed model (b) not only shows better visual qualities, but is also more faithful to the latent noise $z$ such that the chair types with the same latent noise are more similar to each other even at different yaw angles.}}
\label{fig: rc49_gap20}
\end{figure*}

We evaluate the generated chair images using three different metrics. 1) \textbf{Visual quality}: we use Intra-FID \citep{miyato2018cgans, Heusel2017} to measure the distance between real and generated distributions by using features extracted by a pretrained network. 2) \textbf{Label consistency}: average absolute error between the true conditions and the labels predicted by a pretrained network in order to measure whether the generated images are faithful to the given conditions. 3) \textbf{Diversity}: the average entropy of the predicted chair types of the generated images. More details are given in the Supplementary Materials. In summary, our proposed model outperforms all three baselines in terms of visual quality and label consistency, but is slightly weaker at producing diverse images. The result is consistent with the intuition that the proposed penalty term will encourage the model to cover the gaps by generating smoother transitions. However, penalizing the variations of generated images when the conditions are slightly perturbed but the latent noise is kept the same implies that the model is forced to generate similar images when the latent noise is the same regardless of variations in the given conditions. This is demonstrated in Figure \ref{fig: rc49_gap20} which compares the vanilla cGAN and CcGAN against our proposed model. It can be seen that when the latent noise is held the same for each column, the chair types display fewer changes at different yaw angles for the proposed model when compared against the other two models. Although the diversity score is negatively affected, it also suggests that our model has the capability to generate images that are consistent with the latent noises even with different input conditions. If diversity is required for practical purposes, additional regularization terms such as diversity-loss \citep{yang2019diversitysensitive} can be easily incorporated on top of the proposed model.

In addition, while our results are still competitive, we observe that for images, directly computing $||G(\boldsymbol{x},\boldsymbol{z})-G(\boldsymbol{x}+\Delta\boldsymbol{x},\boldsymbol{z})||$ from Equation (\ref{eq: approx_gp}) using pixel-wise distance might not be representative of the underlying semantics of the images with respect to the conditions. An alternative is to project the generator outputs into another metric space where distances between two points are more interpretable - for example, the Riemannian metric on the latent space of variational autoencoders (VAEs) \citep{arvanitidis2018latent, wang2019riemannian}. We will discuss those aspects in more details in future iterations.

\section{Conclusion}
In this work, we provide an attempt to address the issues aroused in training conditional generative adversarial networks (cGANs) when the conditions are continuous and high-dimensional. We propose a simple generator regularization term on the GAN generator loss in the form of Lipschitz penalty. Thus, when the generator is fed with neighboring conditions in the continuous space, the regularization term will leverage the neighbor information and push the generator to generate samples that have similar conditional distributions for each neighboring condition. We demonstrate its robust performance on a range of synthetic and real-world tasks compared to existing methods. Future works include exploring more network structures and integrating other techniques to improve the training of cGANs.

\newpage
\bibliographystyle{informs2014} 
\bibliography{main_paper} 
\ECSwitch


\ECHead{Appendix}

\section{Algorithm for GR-cGAN Training}
We give the algorithms for training a GR-cGAN. If the generator regularization takes the form in Equation \ref{eq: gp_original}, an algorithm for training a GR-cGAN is given in Algorithm \ref{alg:original}. If the generator regularization takes the approximated form in Equation \ref{eq: approx_gp}, please refer to Algorithm \ref{alg:approx}.

\begin{algorithm}[h!] 
\caption{An algorithm for training GR-cGAN with generator regularization as in Equation \ref{eq: gp_original}}
\label{alg:original}
\begin{algorithmic}[1]
\REQUIRE The generator regularization coefficient $\lambda$, the training set $\{\boldsymbol{x}_i, \boldsymbol{y}_i\}_{i=1}^N$,
the batch size $m$, the number of iterations of the discriminator per generator iteration $n$,
Adam hyper-parameters $\alpha$, $\beta_1$ and $\beta_2$, the number of iterations $K$.\\
\REQUIRE $w_{0}$, initial discriminator parameters. $\theta_{0}$, initial generator's parameters.
\FOR{$k=1$ \textbf{to} $K$}\label{alg:2}
    \FOR{$t=1, \ldots, n$}\label{alg:2}
        \STATE Sample a batch of real samples from the training set, denote as $\{\boldsymbol{x}_j,\boldsymbol{y}_j\}^{m}_{j=1}$.\label{alg:3}
        \STATE Sample a batch of random noises independently, $\boldsymbol{z}_j\sim p_z(\boldsymbol{z}),\text{ for }j=1,2,\ldots,m$.\label{alg:4}
        \STATE Discriminator loss $\leftarrow \frac{1}{m}\sum_{j=1}^m\left[\log D(\boldsymbol{x}_j,\boldsymbol{y}_j) + \log(1- D(G(\boldsymbol{x}_j, \boldsymbol{z}_j)))\right]$
        \STATE Update $D$.
    \ENDFOR
    \STATE Sample two batches of real samples  rom the training set independently, denote as $\{\boldsymbol{x}_{j},\boldsymbol{y}_{j}\}^{m}_{j=1}$ and $\{\boldsymbol{x}'_{j},\boldsymbol{y}'_{j}\}^{m}_{j=1}$.
    \STATE Sample a batch of random noises independently, $\boldsymbol{z}_j\sim p_z(\boldsymbol{z})\text{ for }j=1,2,\ldots,m$.
    \STATE Sample random numbers $\epsilon_j\sim U[0,1]$ for $j=1,2,\ldots, m$.
    \STATE $\boldsymbol{x}''_j\leftarrow\epsilon \boldsymbol{x}_j+(1-\epsilon)\boldsymbol{x}’_j$ for $j=1,2,\ldots, m$.
    \STATE $\mathcal{L}_{GR}(G)\leftarrow\frac{1}{m} \sum_{j=1}^m  ||\nabla_{\boldsymbol{x}_j''}G(\boldsymbol{x}_j'',\boldsymbol{z}_j)||$
    \STATE  Generator loss $\leftarrow \frac{1}{m}\sum_{j=1}^m [\log (1-D(G(\boldsymbol{x}_j, \boldsymbol{z}_j)))]+\lambda \mathcal{L}_{G R}(G)$
    \STATE Update $G$.
\ENDFOR
\end{algorithmic}
\end{algorithm}

\begin{algorithm}[h!] 
\caption{An algorithm for training GR-cGAN with generator regularization as in Equation \ref{eq: approx_gp}}
\label{alg:approx}
\begin{algorithmic}[1]
\REQUIRE The generator regularization coefficient $\lambda$, the training set $\{\boldsymbol{x}_i, \boldsymbol{y}_i\}_{i=1}^N$,
the batch size $m$, the number of iterations of the discriminator per generator iteration $n$,
Adam hyper-parameters $\alpha$, $\beta_1$ and $\beta_2$, the number of iterations $K$.\\
\REQUIRE $w_{0}$, initial discriminator parameters. $\theta_{0}$, initial generator's parameters.
\FOR{$k=1$ \textbf{to} $K$}\label{alg:2}
    \FOR{$t=1, \ldots, n$}\label{alg:2}
        \STATE Sample a batch of real samples from the training set, denote as $\{\boldsymbol{x}_j,\boldsymbol{y}_j\}^{m}_{j=1}$.\label{alg:3}
        \STATE Sample a batch of random noises independently, $\boldsymbol{z}_j\sim p_z(\boldsymbol{z}),\text{ for }j=1,2,\ldots,m$.\label{alg:4}
        \STATE Discriminator loss $\leftarrow \frac{1}{m}\sum_{j=1}^m\left[\log D(\boldsymbol{x}_j,\boldsymbol{y}_j) + \log(1- D(G(\boldsymbol{x}_j, \boldsymbol{z}_j)))\right]$
        \STATE Update $D$.
    \ENDFOR
    \STATE Sample two batches of real samples from the training set independently, denote as $\{\boldsymbol{x}_{j},\boldsymbol{y}_{j}\}^{m}_{j=1}$ and $\{\boldsymbol{x}'_{j},\boldsymbol{y}'_{j}\}^{m}_{j=1}$.
    \STATE Sample a batch of random noises independently, $\boldsymbol{z}_j\sim p_z(\boldsymbol{z})\text{ for }j=1,2,\ldots,m$.
    \STATE Sample random numbers $\epsilon_j\sim U[0,1]$ for $j=1,2,\ldots, m$.
    \STATE $\boldsymbol{x}''_j\leftarrow\epsilon \boldsymbol{x}_j+(1-\epsilon)\boldsymbol{x}’_j$ for $j=1,2,\ldots, m$.
    \STATE Sample a batch of perturbations $\Delta \boldsymbol{x}_j\sim p_{\Delta \boldsymbol{x}}(\Delta \boldsymbol{x})\text{ for }j=1,2,\ldots,m$
    \STATE $\mathcal{L}_{\widetilde{GR}}(G)\leftarrow\frac{1}{m}\sum_{j=1}^m[\min(f(\boldsymbol{x}''_j, \Delta \boldsymbol{x}_j, \boldsymbol{z}_j), \tau_1)]$, where $f(\boldsymbol{x}''_j, \Delta \boldsymbol{x}_j, \boldsymbol{z}_j)=\frac{\|G(\boldsymbol{x}_j''+\Delta \boldsymbol{x}_j, \boldsymbol{z}_j)-G(\boldsymbol{x}_j'', \boldsymbol{z}_j)\|}{\|\Delta \boldsymbol{x}_j\|}.$
    \STATE Generator loss $\leftarrow \frac{1}{m}\sum_{j=1}^m [\log (1-D(G(\boldsymbol{x}_j, \boldsymbol{z}_j)))]+\lambda \mathcal{L}_{\widetilde{G R}}(G)$
    \STATE Update $G$.
\ENDFOR
\end{algorithmic}
\end{algorithm}

\begin{figure}[h!]
    \centering                                           
    \includegraphics[width=\columnwidth]{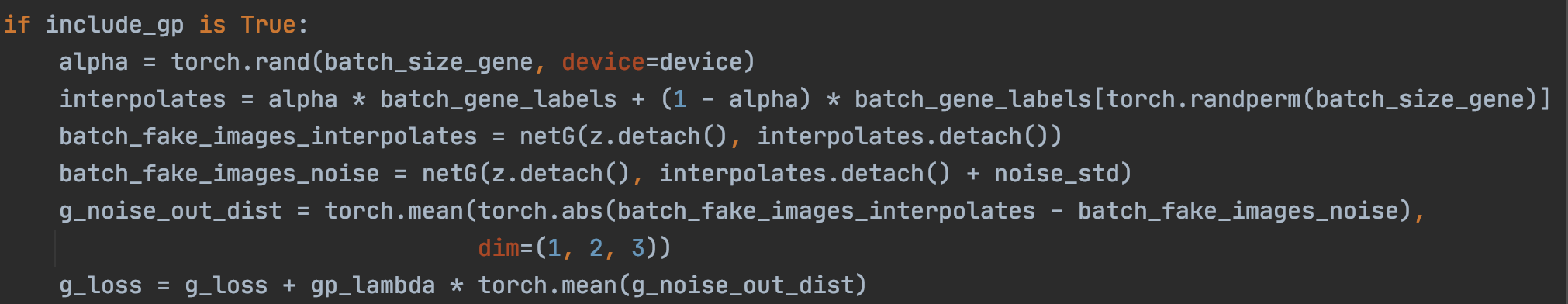}   
    \caption{A sample implementation of the proposed generator regularization in PyTorch.}
    \label{fig: pytorch}
\end{figure}

A sample implementation in PyTorch is shown in Figure \ref{fig: pytorch}.

\newpage
\section{Connection of Generator Regularization to K-Lipschitz Continuous Conditional Distribution}
We formally give the relationship between the conditional distribution learned by a generator and K-Lipschitz continuous conditional distribution aforementioned in Section \ref{sec:analysis} in Theorem \ref{theorem:one}. 
\begin{theorem}
\label{theorem:one}
Suppose that the given two arbitrary conditions $\boldsymbol{x}_1$ and $\boldsymbol{x}_2$, the conditional generator $G$ satisfies
$$
||G(\boldsymbol{x}_1, \boldsymbol{z}_0) - G(\boldsymbol{x}_2, \boldsymbol{z}_0)|| \le K_0\cdot ||\boldsymbol{x}_1-\boldsymbol{x}_2||
$$
for any fixed $\boldsymbol{z}_0$. We have 
$$
W(G(\boldsymbol{x}_1,\boldsymbol{ z}), G(\boldsymbol{x}_2, \boldsymbol{z}))\le K_0\cdot ||\boldsymbol{x}_1-\boldsymbol{x}_2||,
$$
where $\boldsymbol{z}\sim p_{\boldsymbol{z}}(\boldsymbol{z})$.
\end{theorem}
We prove Theorem \ref{theorem:one} using the following Lemma \ref{lemma:dist_func_rv}.
\begin{lemma}
\label{lemma:dist_func_rv}
Denote the support of a random variable $Z$ as $R_Z$. Functions $f$ and $g$ are defined on $R_Z$. Denote the distribution of $f(Z)$ and $g(Z)$ as $\mathcal{P}_f(Z)$ and $\mathcal{P}_g(Z)$ respectively. If we have $\max_{z\in R_Z}\left\|f(z)-g(z)\right\|\le K$, then the Wasserstein distance between $\mathcal{P}_f(Z)$ and $\mathcal{P}_g(Z)$ satisfies $W(\mathcal{P}_f(Z),\mathcal{P}_g(Z))\le K$.
\end{lemma}

\begin{proof}
Denote $X=f(Z)$ and $Y=g(Z)$, and the distribution of $f(Z)$ and $g(Z)$ as $\mathcal{P}_f(Z)$ and $\mathcal{P}_g(Z)$ respectively. Clearly, 
$$
\mathcal{P}_f(x)=\int_{z:f(z)=x,z\in R_Z}\mathcal{P}_Z(z)\,dz
$$
and 
$$
\mathcal{P}_g(y)=\int_{z:g(z)=y,z\in R_Z}\mathcal{P}_Z(z)\,dz.
$$
The support of $f(Z)$ and $g(Z)$, i.e., the set $\{f(z):z\in R_Z\}$ and $\{f(z):z\in R_z\}$ is denoted as $f(R_Z)$ and $g(R_Z)$. Define a joint distribution of $X$ and $Y$ as
\begin{align}
\gamma_{0}(x, y)=\left\{\begin{array}{ll}\int_{z: f(z)=x \text { and } g(z)=y} \mathcal{P}(z) dz \quad & z \in R_{Z}, \text { s.t. } f(z)=x \text { and } g(z)=y \\ 0 & \text { o.w. }\end{array}\right.
\end{align}
$\gamma_0$ is intentionally designed such that the marginal distribution of $X$ and $Y$ is precisely $\mathcal{P}_f(X)$ and $\mathcal{P}_g(Y)$:
\begin{align}
\begin{split}
\int_{x\in f(R_Z)} \gamma_0(x,y)\,dx&=\int_{x\in f(R_Z)}\int_{z:f(z)=x \text{ and } g(z)=y}\mathcal{P}_Z(z)\,dz\,dx
\\&=\int_{z:g(z)=y,z\in R_Z}\mathcal{P}_z(z)\,dz
\\&=\mathcal{P}_g(y)
\end{split}
\end{align}
and
\begin{align}
\begin{split}
\int_{y\in g(R_Z)} \gamma_0(x,y)\,dy&=\int_{y\in g(R_Z)}\int_{z:f(z)=x \text{ and } g(z)=y}\mathcal{P}_Z(z)\,dz\,dy
\\&=\int_{z:f(z)=x,z\in R_Z}\mathcal{P}_z(z)\,dz
\\&=\mathcal{P}_f(x).
\end{split}
\end{align}
By the definition of Wasserstein distance, 
\begin{align*}
\begin{split}
W(\mathcal{P}_f(Z),\mathcal{P}_g(Z))&=\inf _{\gamma \in \Pi\left(\mathcal{P}_{f}, \mathcal{P}_{g}\right)} \mathbb{E}_{(x, y) \sim \gamma}[\|x-y\|]\\
&\le \mathbb{E}_{(x,y)\sim \gamma_0}[\|x-y\|]\\
&=\int_{x\in f(R_Z)}\int_{y\in g(R_Z)}\gamma_0(x,y)\cdot\|x-y\|\,dx\,dy\\
&=\int_{x\in f(R_Z)}\int_{y\in g(R_Z)}\int_{z:f(z)=x \text{ and } g(z)=y}\mathcal{P}(z)\cdot\|x-y\|\,dz\,dx\,dy\\
&=\int_{z\in R_Z}\mathcal{P}_{z}(z)\cdot\|f(z)-g(z)\|\,dz\\
&\le \int_{z\in R_Z}\mathcal{P}_Z(z)\cdot K\,dz\\
&=K.
\end{split}
\end{align*}
\end{proof}
The proof of Theorem \ref{theorem:one} is obvious using Lemma \ref{lemma:dist_func_rv}. In Theorem \ref{theorem:one}, given the fixed $\boldsymbol{x}_1$ (or $\boldsymbol{x}_2$), the generator $G$ can be viewed as a function $G(\boldsymbol{x}_1,\cdot)$ (or $G(\boldsymbol{x}_2, \cdot)$) that maps a random noise $\boldsymbol{z}$ to $G(\boldsymbol{x}_1,\boldsymbol{z})$ (or $G(\boldsymbol{x}_2,\boldsymbol{z})$). Take the random variable $Z$ in Lemma \ref{lemma:dist_func_rv} as $\boldsymbol{z}$. Take $G(\boldsymbol{x}_1,\cdot)$ and $G(\boldsymbol{x}_2, \cdot)$ as the functions $f$ and $g$ in Lemma \ref{theorem:one}. Then Theorem \ref{theorem:one} is evident.

\section{More Details of the Experiments}
\subsection{Circular 2-D Gaussians}
\textbf{Network architectures:} We use the same network architecture setting as in \cite{ding2021ccgan}. Please refer to Table \ref{tab:cir} for details.

\begin{table}[ht!]
\centering 
\subfigure[Generator]
{
\begin{minipage}{0.45\columnwidth}
\label{tab:cir_generator}
\centering                                           
\begin{tabular}{c}
\toprule
$\boldsymbol{z} \in \mathbb{R}^{2} \sim N(0, I) ; \boldsymbol{y} \in \mathbb{R}$\\
\midrule
$\operatorname{concat}(\boldsymbol{z}, \sin (\boldsymbol{y}), \cos (\boldsymbol{y})) \in \mathbb{R}^{4}$\\
\midrule
$\mathrm{fc} \rightarrow 100 ; \mathrm{BN} ; \operatorname{ReLU}$\\
\midrule
$\mathrm{fc} \rightarrow 100 ; \mathrm{BN} ; \operatorname{ReLU}$\\
\midrule
$\mathrm{fc} \rightarrow 100 ; \mathrm{BN} ; \operatorname{ReLU}$\\
\midrule
$\mathrm{fc} \rightarrow 100 ; \mathrm{BN} ; \operatorname{ReLU}$\\
\midrule
$\mathrm{fc} \rightarrow 100 ; \mathrm{BN} ; \operatorname{ReLU}$\\
\midrule
$\mathrm{fc} \rightarrow 100 ; \mathrm{BN} ; \operatorname{ReLU}$\\
\midrule
$\mathrm{fc} \rightarrow 2$\\
\bottomrule
\end{tabular}
\end{minipage}
}
\subfigure[Discriminator]
{
\begin{minipage}{0.45\columnwidth}
\label{tab:cir_disc}
\vspace{0.1in}
\centering                                           
\begin{tabular}{c}
\toprule
$\text { A sample } \boldsymbol{x} \in \mathbb{R}^{2} \text { with label } \boldsymbol{y} \in \mathbb{R}$\\
\midrule
$\operatorname{concat}(x, \sin (y), \cos (y)) \in \mathbb{R}^{4}$\\
\midrule
$\mathrm{fc} \rightarrow 100 ; \text { ReLU }$\\
\midrule
$\mathrm{fc} \rightarrow 100 ; \text { ReLU }$\\
\midrule
$\mathrm{fc} \rightarrow 100 ; \text { ReLU }$\\
\midrule
$\mathrm{fc} \rightarrow 100 ; \text { ReLU }$\\
\midrule
$\mathrm{fc} \rightarrow 100 ; \text { ReLU }$\\
\midrule
$\mathrm{fc} \rightarrow 1 ; \text { Sigmoid }$\\
\bottomrule
\end{tabular}
\end{minipage}}
\caption{Network architectures for the generator and discriminator of the experiments in Section \ref{sec:cir_2d}. \textmd{“fc” represents a fully-connected layer. “BN” denotes batch normalization. The label $\boldsymbol{y}$ is treated as a real scalar so its dimension is $1 .$ We use $\boldsymbol{y}$, $\sin (\boldsymbol{y})$ and $\cos(\boldsymbol{y})$ together as the input to the generator networks.}}
\label{tab:cir}
\end{table}

\textbf{Training steps:} The training steps is also the same as in \cite{ding2021ccgan}. All GANs are trained for 6000 iterations on the training set with the Adam (Kingma \& $\mathrm{Ba}, 2015$ ) optimizer (with $\beta_{1}=0.5$ and $\beta_{2}=0.999$ ), a constant learning rate $5 \times 10^{-5}$ and batch size $128 .$ The hyper parameters of CcGAN takes the same value as in Section S.VI.B of \cite{ding2021ccgan}. The $\lambda$ of GR-cGAN is set to 0.02, with the generator regularization term computed by Equation \ref{eq: gp_original}.

\subsection{Multivariate Gaussian}
\textbf{Dataset:} In the experiments in Section \ref{sec:mul_gaussian}, the training data are sampled from a multivaraite Gaussian distribution $\mathcal{N}(\boldsymbol{\mu}, \boldsymbol{\Sigma})$. The parameters of this Gaussian, $\boldsymbol{\mu}$ and $\boldsymbol{\Sigma}$, are pre-specified in the following steps. Each element of $\boldsymbol{\mu}$ is randomly drawn from $U[10,15]$. The covariance matrix $\boldsymbol{\Sigma}$ is specified such that $\boldsymbol{\Sigma}$ is positive semi-definite and each element of $\boldsymbol{\Sigma}$ takes value in range $[-0.25, 0.25]$. For the specific details, please refer to our code.

\begin{table}[ht!]
\centering 
\subfigure[Generator]
{
\begin{minipage}{0.45\columnwidth}
\label{tab:cir_generator}
\vspace{0.1in}
\centering                                           
\begin{tabular}{c}
\toprule
$\boldsymbol{z} \in \mathbb{R}^{k-p} \sim N(0, I) ; \boldsymbol{x} \in \mathbb{R}^{p}$\\
\midrule
$\mathrm{fc} \rightarrow 512 ; \operatorname{LeakyReLU}$\\
\midrule
$\mathrm{fc} \rightarrow 512 ; \operatorname{LeakyReLU}$\\
\midrule
$\mathrm{fc} \rightarrow 512 ; \operatorname{LeakyReLU}$\\
\midrule
$\mathrm{fc} \rightarrow k-p$\\
\bottomrule
\end{tabular}
\end{minipage}
}
\subfigure[Discriminator]
{
\begin{minipage}{0.45\columnwidth}
\label{tab:cir_disc}
\centering                                           
\begin{tabular}{c}
\toprule
$\text { A sample } \boldsymbol{y} \in \mathbb{R}^{k-p} \text { with label } \boldsymbol{x} \in \mathbb{R}^p$\\
\midrule
$\operatorname{concat}(\boldsymbol{x}, \boldsymbol{y}) \in \mathbb{R}^{k}$\\
\midrule
$\mathrm{fc} \rightarrow 512 ; \operatorname{LeakyReLU}$\\
\midrule
$\mathrm{fc} \rightarrow 512 ; \operatorname{LeakyReLU}$\\
\midrule
$\mathrm{fc} \rightarrow 512 ; \operatorname{LeakyReLU}$\\
\midrule
$\mathrm{fc} \rightarrow 1$\\
\bottomrule
\end{tabular}
\end{minipage}}
\caption{Network architectures for the generator and discriminator of the experiments in Section \ref{sec:mul_gaussian}. \textmd{“fc” represents a fully-connected layer. The label $\boldsymbol{y}$ is a vector of dimension is $k-p$. The noise $\boldsymbol{z}$ is set to a vector of the same dimension as $\boldsymbol{y}$. We set the negative slope of LeakyReLU to 0.1.}}
\label{tab:mul_gaussian}
\end{table}

\textbf{Network architectures and training steps:} We use a fully connected neural network for both the generator and discriminator. Given the dimension of $\boldsymbol{x}$ as $p$, and the dimension of $\boldsymbol{y}$ as $k-p$, the generator and discriminator are given as in Table \ref{tab:mul_gaussian}. We use a Wasserstein type discriminator with gradient penalty as in \cite{gulrajani2017improved}. The gradient penalty coefficient is set to 0.1 in all the experiments. An Adam optimizer with $\beta_1=0.5$,$\beta_2=0.9$ and learning rate $2\times 10^{-5}$ is applied. The batch size is set to 256. To compute the generator regularization, we adopt the approximated form as in \ref{eq: approx_gp} with $\lambda=1$. The distribution of the perturbation term $p_{\Delta \boldsymbol{x}}(\Delta \boldsymbol{x})$ is implicitly defined by uniformly sampling $\Delta \boldsymbol{x}$ on the surface of a $p$-dimensional ball with radius 0.1. (We also test setting $p_{\Delta \boldsymbol{x}}(\Delta \boldsymbol{x})$ as a multivariate Gaussian distribution and find
it is numerically unstable because the denominator in \ref{eq: approx_gp} can be arbitrarily small.) The term $\tau_1$ in Equation \ref{eq: approx_gp} is set to $+\infty$. All the models are trained for 50,000 iterations.

\begin{figure}[ht!]
\centering 
\subfigure[cGAN]
{
\begin{minipage}{0.9\columnwidth}
\centering                                           
\includegraphics[width=\columnwidth]{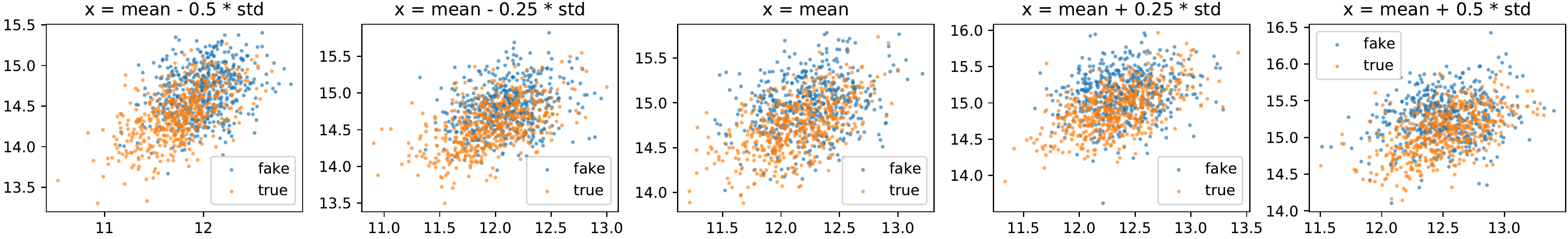}   
\end{minipage}
}
\subfigure[GR-cGAN]
{
\begin{minipage}{0.9\columnwidth}
\centering                                           
\includegraphics[width=\columnwidth]{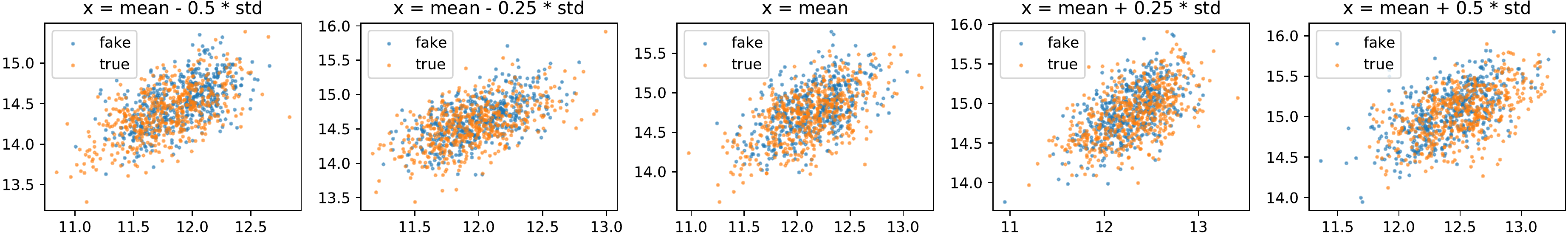}   
\end{minipage}
}
\caption{Visual results of the multivariate Gaussian experiment on more labels. \textmd{Given each label, we use cGAN and GR-cGAN to generate 250 fake samples, and plot them in orange dots. Besides, we sample 250 points from the true conditional distribution and plot them in blue dots. Compared with (a), the distribution of true and false samples in (b) is closer.}}
\label{fig:highdim_more}
\end{figure}

\textbf{More results}
Given the parameters of the multivariate Gaussian distribution $\boldsymbol{\mu}$ and $\boldsymbol{\Sigma}$, we compute the marginal standard deviation of each dimension as $\boldsymbol{\sigma}$. Denote the first $p$ dimensions of $\boldsymbol{\sigma}$ as $\boldsymbol{\sigma}_{1:p}$. We set the label $\boldsymbol{x}$ to $\boldsymbol{\mu}_{1:p}-0.5\boldsymbol{\sigma}_{1:p}$, $\boldsymbol{\mu}_{1:p}-0.25\boldsymbol{\sigma}_{1:p}$, $\boldsymbol{\mu}_{1:p}$, $\boldsymbol{\mu}_{1:p}+0.25\boldsymbol{\sigma}_{1:p}$ and $\boldsymbol{\mu}_{1:p}+0.5\boldsymbol{\sigma}_{1:p}$. Given each label, we use the same steps as described in Section \ref{sec:multi_gaussian_result} to get 250 true samples and fake samples. These samples are plotted in Figure \ref{fig:highdim_more}. The conditional distribution given by GR-cGAN is closer to the true conditional distribution, compared to using cGAN.

\begin{figure}[ht!]
\begin{center}
\centerline{\includegraphics[width=0.3\columnwidth]{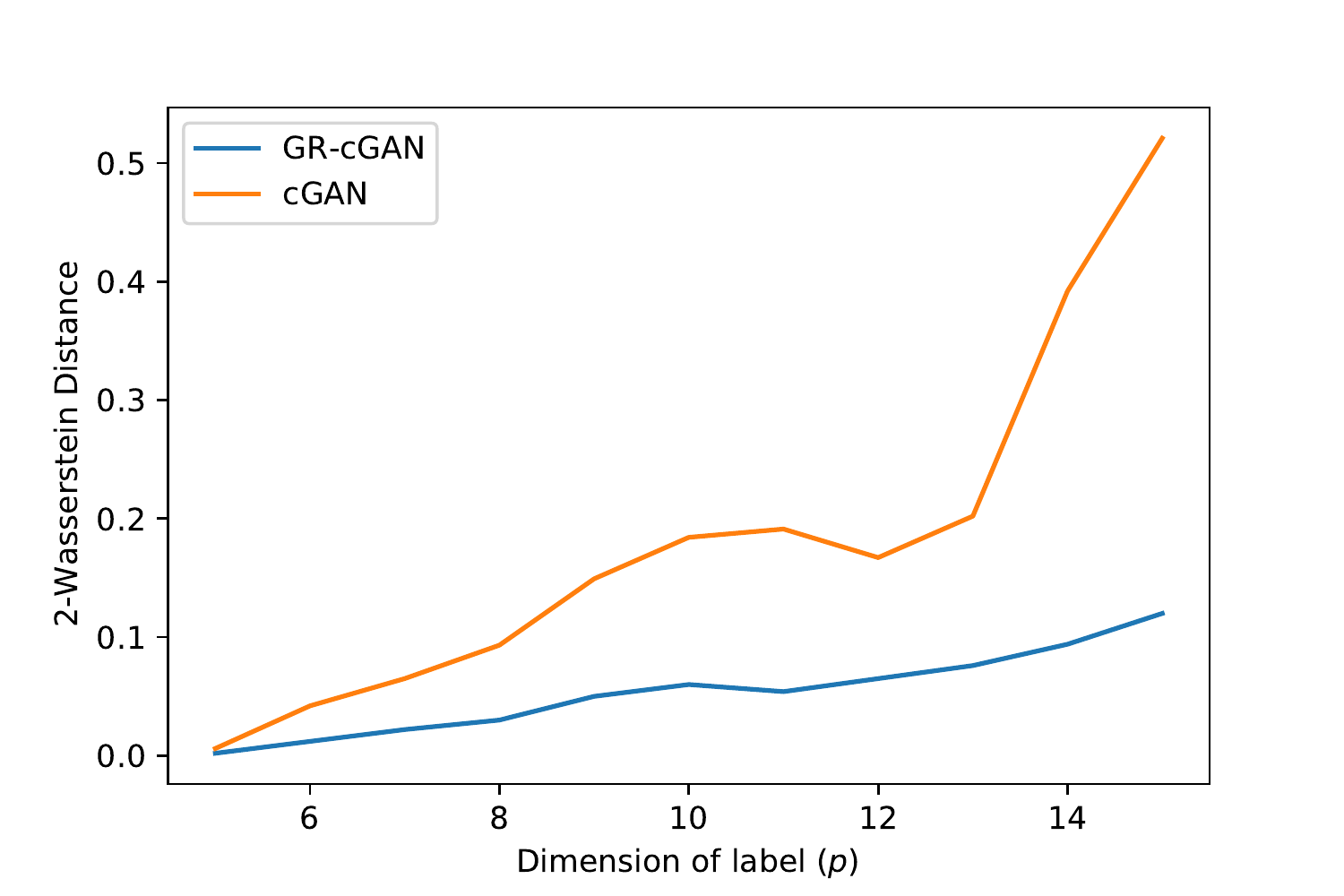}}
\caption{The 2-Wasserstein Distance between the conditional distribution obtained by generator and the real conditional distribution. }
\label{fig:highdim_dist}
\end{center}
\end{figure}

We also give numerical evaluations. We generate 100 labels from the distribution $\mathcal{N}(\boldsymbol{\mu}, \boldsymbol{\Sigma})$. For each label, we calculate the 2-Wasserstein Distance between true and fake samples. This value is used to roughly measure the distance between the conditional distribution obtained by generator and the real conditional distribution. We average the distances obtained on 100 labels. We set the dimension $p$ from 5 to 15 (while keeping $k$ as $p+2$) and present the results in Figure \ref{fig:highdim_dist}. GR-cGAN outperforms cGAN on each dimension setting.

\subsection{RC-49}
\textbf{Dataset:} RC-49 \citep{ding2021ccgan} is a synthetic dataset created by rendering 49 different types of 3-D chair models at different yaw angles in order to evaluate the performance of GANs on continuous and scalar regression labels. The image for each chair type is collected from 0.1 to 89.9 degrees with 0.1 degree increments, which results in a total of 44,051 RGB images each of size $64 \times 64$. At each angle, we randomly select 25 images for training. 

\textbf{Evaluation: } At evaluation, each model is asked to generate 200 fake images at each of the 899 distinct angles. We pretrain three models which are then used to evaluate the GAN models - an autoencoder with a latent dimension of 512, a regression-oriented ResNet-34 \citep{He2015} and a classification-oriented ResNet-34 \citep{He2015} using all 44,051 images. The autoencoder is trained to reconstruct the images under Mean Squared Error. The regression-oriented ResNet-34 is trained to predict the angle at which a given image is taken. The classification-oriented ResNet-34 is trained to predict a given image belongs to which of the 49 chair types. All three models are trained for 200 epochs with a batch size of 256. Using those three models, we produce the following metrics:
\begin{itemize}
    \item Intra-FID \citep{miyato2018spectral, Heusel2017}: At each of the 899 angles, we compute the FID between 49 real images and 200 fake images in terms of the
latent vector of the pre-trained autoencoder. The final Intra-FID score is the mean FID score over all angles.
    \item Diversity: At each evaluation angle, the classification-oriented ResNet-34 is used to predict which of the 49 chair types a given image belongs to for each of the 200 fake images. We calculate the entropy from predicted chair types and report the average of over all angles.
    \item Label Score: The regression-oriented ResNet-34 is used to predict which of the 899 angles a fake image belongs to. We then take the mean absolute distance between the predicted angles and the assigned angles over all fake images.
\end{itemize}
    
\textbf{Task: } In the original setting, training is done using images at angles where the last digit is odd, which means that training samples are given with gaps of 0.2 degree. However, in real-world settings we often observe larger gaps in the training data. Thus, we increase the gap to $\{5, 10, 18, 30\}$ in order to better evaluate the robustness of the GAN models. We also observed that test angles that lie outside of the training angles (for example, test angle at 1 degree if the minimum train angle is 5 degrees) usually yield suboptimal performances. Therefore, we first divide all angles evenly into groups of size $\{5, 10, 18, 30\}$ and always use the middle $50\%$ as test and the outer $50\%$ as train. For instance, for gap $=30$, we let angles $(< 7.5)$ degrees and between $(22.5, 37.5)$ degrees as train and between $(7.5, 22.5)$ as test and so on.

\textbf{Network Architecture: } We adopted the label embedding network for all the GAN models for fair comparison. We first train an encoder network that predicts its condition given the image. The latent vector of the second last layer is extracted as the hidden representation of the image. We then train a second network that tries to predict the hidden representation given the label. Note that gaussian noise is added to the input of the second network to increase its converge. More details can be found in \cite{ding2021ccgan}.

\end{document}